\documentclass[10pt,twocolumn,letterpaper]{article}
\usepackage{cvpr}
\usepackage{graphicx}
\usepackage{amsmath}
\usepackage{amssymb}
\usepackage{booktabs}
\usepackage{multirow}
\usepackage{placeins}
\usepackage{graphics,epsfig,color}
\usepackage[pagebackref,breaklinks,colorlinks]{hyperref}

\usepackage[capitalize]{cleveref}
\crefname{section}{Sec.}{Secs.}
\Crefname{section}{Section}{Sections}
\Crefname{table}{Table}{Tables}
\crefname{table}{Tab.}{Tabs.}

\def\cvprPaperID{770} 
\def\confName{CVPR}
\def\confYear{2022}

\begin{document}

\title{Putting People in their Place: Monocular Regression of 3D People in Depth}
\makeatletter
\g@addto@macro\@maketitle{
\centering
  \includegraphics[width=0.94\textwidth]{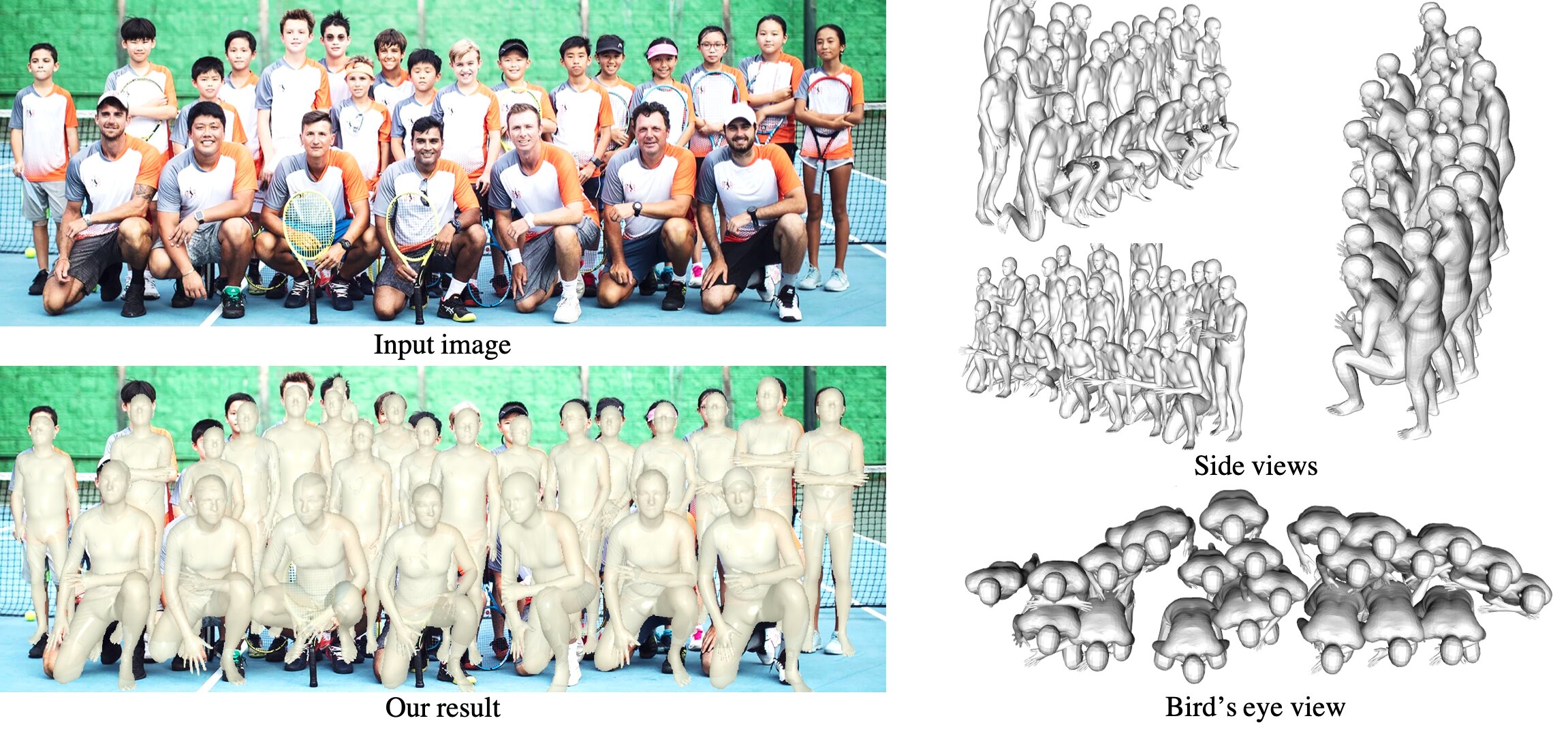}\vspace{-4mm}
\captionof{figure}{\textbf{Monocular reconstruction of multiple 3D people with coherent depth reasoning.} We introduce BEV, a monocular one-stage method with an efficient new ``bird's-eye-view" representation that enables the network to explicitly reason about people in 3D. }\vspace{5mm} 
\label{fig:teaser}
}

\author{Yu Sun$^1$\thanks{This work was done when Yu Sun was an intern at Explore Academy of JD.com. }\quad
Wu Liu$^{2}$\thanks{Corresponding author.} \quad
Qian Bao$^2$ \quad
Yili Fu$^{1\dagger}$\quad
Tao Mei$^2$\quad
Michael J. Black$^3$ \quad \\

$^1$Harbin Institute of Technology, Harbin, China \quad $^2$ Explore Academy of JD.com, Beijing, China \\
$^3$Max Planck Institute for Intelligent Systems, T\"ubingen, Germany\\
{\tt\small \texttt{yusun@stu.hit.edu.cn, liuwu1@jd.com, baoqian@jd.com, meylfu@hit.edu.cn}}\\
{\tt\small\texttt{tmei@jd.com, black@tuebingen.mpg.de}}\vspace{-2mm}
}

\makeatother

\maketitle

\begin{abstract}
Given an image with multiple people, our goal is to directly regress the pose and shape of all the people as well as their relative depth. Inferring the depth of a person in an image, however, is fundamentally ambiguous without knowing their height. This is particularly problematic when the scene contains people of very different sizes, e.g. from infants to adults. To solve this, we need several things. First, we develop a novel method to infer the poses and depth of multiple people in a single image. While previous work that estimates multiple people does so by reasoning in the image plane, our method, called BEV, adds an additional {\em imaginary} Bird's-Eye-View representation to explicitly reason about depth. BEV reasons simultaneously about body centers in the image and in depth and, by combing these, estimates 3D body position. Unlike prior work, BEV is a single-shot method that is end-to-end differentiable. Second, height varies with age, making it impossible to resolve depth without also estimating the age of people in the image. To do so, we exploit a 3D body model space that lets BEV infer shapes from infants to adults. Third, to train BEV, we need a new dataset. Specifically, we create a ``Relative Human" (RH) dataset that includes age labels and relative depth relationships between the people in the images. Extensive experiments on RH and AGORA demonstrate the effectiveness of the model and training scheme. BEV outperforms existing methods on depth reasoning, child shape estimation, and robustness to occlusion. The code\footnote{\url{https://github.com/Arthur151/ROMP}} and dataset\footnote{\url{https://github.com/Arthur151/Relative_Human}} are released for research purposes.
\end{abstract}

\section{Introduction}
\label{sec:intro}

In this article, we focus on simultaneously estimating the 3D pose and shape of all people in an RGB image along with their relative depth.
There has been rapid progress \cite{liu2022recent} on regressing the 3D pose and shape of individual (cropped) people \cite{kocabas2020vibe,hmr,sun2019dsd-satn,kolotouros2019spin,zhou2018unsupervised,zhang2021pymaf,Kocabas_SPEC_2021,moon2020pose2pose,pavlakos2019texturepose,keep,Zeng_2020_CVPR,zhang2021lightweight} as well as the direct regression of groups \cite{romp,jiang2020coherent}.
Neither class of methods explicitly reasons about the depth of people in the scene.
Such depth reasoning is critical to enable a deeper understanding of the scene and the multi-person interactions within it.
To address this, we propose a unified method that jointly regresses multiple people and their relative depth relations in one shot from an RGB image.

While previous multi-person methods perform well in constrained experimental settings, they struggle with severe occlusion, diverse body size and appearance, the ambiguity of monocular depth, and in-the-wild cases \cite{jiang2020coherent,moon2019camera,zhen2020smap,wang2020hmor}.
These challenges lead to unsatisfactory performance in crowded scenes, including detection misses, similar predictions for overlapping people, and all predictions having a similar height. 
We observe two inter-related limitations that result in these failures.
First, the architecture of the regression networks is closely tied to the 2D image, while the people actually inhabit 3D space. 
We address this with a new architecture that reasons in 3D.
Second, depth estimation is fundamentally ambiguous due to the unknown height of the people in the image and it is difficult to obtain training data of images with ground-truth height and depth.
To address this, we present a new dataset and novel losses that allow training without having metric depth.


We observe that crowded scenes contain rich information about the relative relationships between people, which can be exploited for both training and validation of depth reasoning.
However, we still lack a powerful representations to learn from these cases.
A few learning-based methods have been proposed for reasoning about the depth of predicted body meshes~\cite{jiang2020coherent} or 3D poses \cite{moon2019camera,zhen2020smap,wang2020hmor}.
Unfortunately, they all reason about depth via 2D representations, such as RoI-aligned features \cite{moon2019camera,jiang2020coherent} or a 2D depth map \cite{zhen2020smap,wang2020hmor}.
These regression-based 2D representations have inherent drawbacks for representing the 3D world.
The lack of an explicit 3D representation in the networks makes it challenging for these methods to deal with crowded scenes in which people overlap at different depths.
Therefore, we argue that an explicit 3D representation is needed.

\begin{figure*}[t]
	\centerline{\includegraphics[width=1.00\textwidth]{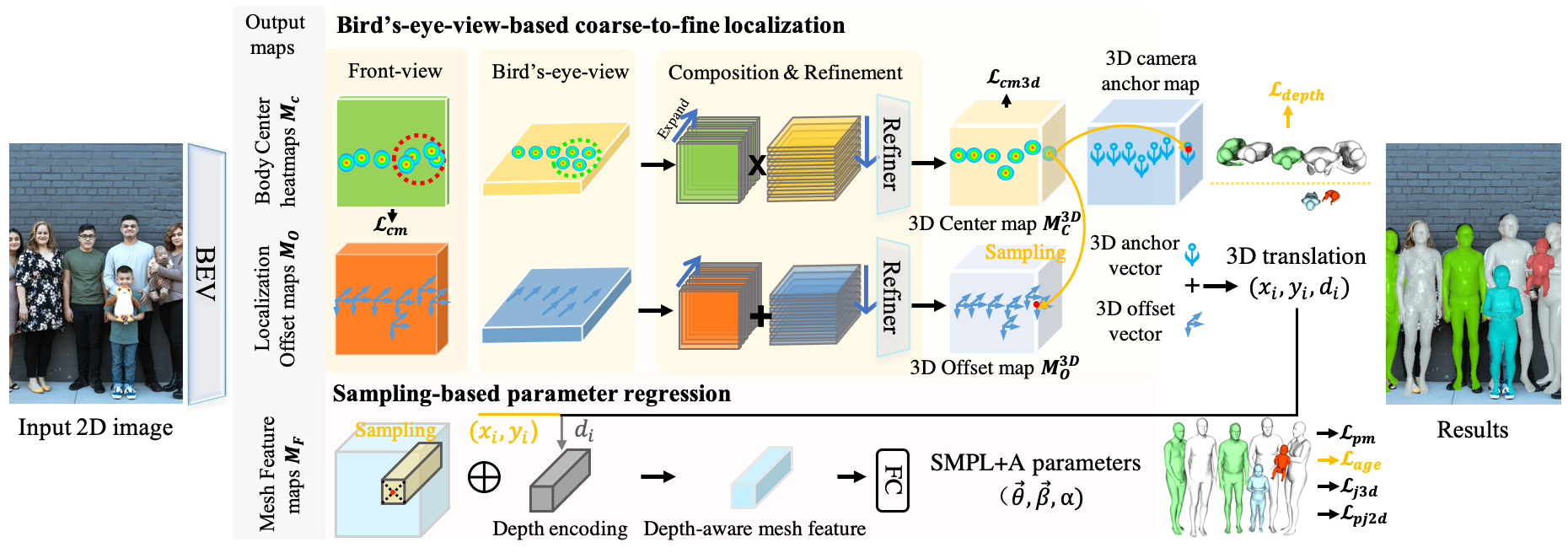}}
	\vspace{-3mm}
	\caption{Overview. Given an RGB image, BEV first estimates the 3D translation of all people in the scene via compositing the front-view and the bird's-eye-view predictions. Then guided by the 3D translation, we sample the mesh feature of each person to regress their age-aware SMPL+A parameters. See Sec.~\ref{sec:overview} for details.}
	\label{fig:framework}
\end{figure*}

To achieve this, we develop BEV (for Bird's Eye View), a unified one-stage method for monocular reconstruction and depth reasoning of multiple 3D people.
We take inspiration from ROMP~\cite{romp}, a one-stage, multi-person, regression method that directly estimates multiple 2D front-view maps for 2D human detection, positioning, and mesh parameter regression without depth reasoning.
With ROMP, the network can only reason about the 2D location of people in the image plane.
To go beyond this, we need to enable the network to efficiently reason about depth as well.
To that end, we introduce a new {\em imaginary} 2D ``bird's-eye-view" map that represents the likely centers of bodies in depth. 
To be clear, BEV takes only a single 2D image; the overhead view is inferred, not observed.
BEV uses a powerful and efficient localization pipeline, performing bird's-eye-view-based coarse detection and fine localization in parallel.
We employ the 2D heatmaps for coarse detection from both the front (image) and bird's eye views.
BEV combines these heatmaps to obtain a 3D heatmap, as illustrated in Fig.~\ref{fig:framework}.
By learning the front and the bird's-eye view together, BEV explicitly models how people appear in images and in depth.
This enables BEV to learn from available 2D and 3D annotations. 
BEV also uses a novel 3D Offset map to refine the initial coarse detections.
From these coarse and fine maps, we obtain the 3D translation of all people in the scene.
BEV transforms these predictions from the latent 3D Center-map space to an explicit camera-centric 3D space.
Given these 3D translation predictions, BEV samples the features of all the people from a predicted mesh feature map and regresses the final SMPL \cite{smpl} parameters.
Distinguishing people at different depths enables BEV to estimate multiple people even with severe occlusion as illustrated in Fig.~\ref{fig:teaser}.

Even with a powerful 3D representation, we need an appropriate training scheme to ensure generalization. The main reason is that without knowing subject height, we lack effective constraints to alleviate the depth/height ambiguity under perspective projection. In particular, height varies with age, making it impossible to resolve depth without also estimating the age of people in the image. The ambiguity causes incorrect depth estimates for children and infants, limiting the generalization of existing methods. 
Unfortunately, existing 3D datasets with multiple people have limited diversity in height and age, so they cannot be used to improve or evaluate generalization.

Since collecting ground-truth 3D data in the wild is difficult, we instead train BEV using cost-effective weak labels of in-the-wild images. 
Specifically, we collect a dataset, named ``Relative Human" (RH), that contains weak annotations of {\em depth layers} and human ages categorized into the groups adult, teenager, child, and infant.
Moreover, we propose a weakly supervised training scheme (WST) to effectively learn from these weak supervision signals.
For instance, we use a piece-wise loss function that exploits the depth layers to penalize incorrect relative depth orders.
Exploiting age information to constrain height is tricky.
While age and height are correlated, heights can vary significantly within the same age group.
Consequently, we develop an ambiguity-compatible mixed loss function that encourages body shapes with heights that lie within an appropriate range for each age group.

We evaluate BEV on three multi-person datasets: in-the-wild using the 2D RH dataset and in 3D using the real CMU Panoptic \cite{cmu_panoptic} and the synthetic AGORA \cite{patel2021agora} datasets.
On RH, compared with previous methods \cite{moon2019camera,jiang2020coherent,zhen2020smap,wang2020hmor}, BEV is more accurate in relative depth reasoning and pose estimation. 
On CMU Panoptic, BEV outperforms previous methods~\cite{zanfir2018deep,zanfir2018monocular,jiang2020coherent,romp,choi20223dcrowdnet} in 3D pose estimation.
On AGORA, BEV significantly improves detection and achieves state-of-the-art results on ``AGORA kids" in terms of the mesh reconstruction error.
Also, fine-tunning on RH in a weakly supervised manner significantly improves the results for all age groups, especially for young people.

In summary, the main contributions are: (1)  We construct a 3D representation to alleviate the monocular depth ambiguity via combining a front-view representation with an imaginary bird's eye view.
(2) We collect the Relative Human dataset with weak annotations of in-the-wild images, which facilitates the training and evaluation on monocular depth reasoning in multi-person scenes.
(3) We develop a weakly supervised training scheme to learn from weak depth annotations and to exploit age information.

\section{Related Work}
\label{sec:related_work}
\textbf{Monocular 3D mesh regression from natural scenes. }
Here, we focus on regressing a 3D body mesh using a parametric model like SMPL from a single RGB image. 
Most methods can be divided into multi-stage or single-stage approaches. 
For general multi-person cases, most existing methods~\cite{keep,hmr,pavlakos2019texturepose,kolotouros2019spin,moon2020pose2pose} are based on a typical two-stage framework, which first detects people and then estimates the parameters of each person separately.
Recent methods focus on exploring various supervision~\cite{rong2019delving} signals, such as temporal coherence~\cite{kocabas2020vibe}, contour alignment~\cite{xiu2022icon,Dwivedi_DSR_2021,pavlakos2018learning},  self-contact~\cite{muller2021self}, ground constraints~\cite{rempe2021humor,yi2022mover}, or global human trajectory~\cite{yuan2022glamr} to enhance the geometric/dynamic consistency. 
However, for depth reasoning about all people in the scene, these multi-stage methods are not ideal.
The processing of individual cropped people cannot exploit the scene context or reason about depth ordering. 

A few one-stage methods~\cite{romp,mehta2018single} estimate multiple 3D people simultaneously. 
Given a single image, ROMP \cite{romp} outputs a 2D Body Center Heatmap, Camera Map, and Parameter Map for 2D human detection, positioning, and mesh parameter regression, respectively.
At the position parsed from the 2D Body Center heatmap, ROMP samples the final mesh parameters from the Camera and Parameter maps.
These one-stage methods enjoy a holistic view of the image, which is more suitable for depth reasoning.
However, they are based on 2D representations that do not represent depth.
Like most methods, they model adults (with SMPL),  train on images of adults, and therefore only predict adults.
To tackle the limitations of their 2D representation and age bias, we propose BEV and its training scheme of learning age priors that constrain body height. 

\textbf{Monocular depth reasoning. } 
Most previous methods place bodies in depth via post-processing.
Due to their 2D-based pipeline and lack of height prior for different age groups, their results are unsatisfying.
A few learning-based methods, like 3DMPPE~\cite{moon2019camera} and CRMH~\cite{jiang2020coherent}, address multi-stage depth reasoning. 
3DMPPE uses image features to refine the bounding-box-based depth predictions.
CRMH learns from instance segmentation to distinguish the relative depth between overlapping people. 
However, instance segmentation is expensive and unable to promote the learning of depth relations in cases without overlapping.
SMAP~\cite{zhen2020smap} and HMOR~\cite{wang2020hmor} employ a 2D depth map to represent the root depth of 3D pose at each pixel. 
However, in crowded scenes, these 2D representations are ambiguous.
In contrast, BEV adopts a novel bird's-eye-view-based 3D representation to distinguish people at different depths, therefore, it is more robust to the overlapping cases.
Most recently, Ugrinovic et al.~\cite{ugrinovic2021body} propose an optimization-based method to refine the 3D translation of estimated body meshes.
They fit the 3D body mesh to the detected 2D poses and force the feet to touch the ground. 
In contrast, our learning-based, one-stage, framework is more efficient and flexible, and can adapt to more scenarios, such as jumping.
Albiero et al.~\cite{albiero2021img2pose} estimate the depth of all faces in a crowd in one shot by regressing their 6DoF pose; they do not deal with shape variation or articulation.


\section{Method}
\label{sec:method}

\subsection{Overview}\label{sec:overview}

The overall framework is illustrated in Fig.~\ref{fig:framework}.
BEV adopts a multi-head architecture. 
Given a single RGB image as input, BEV outputs 5 maps.
For coarse-to-fine localization, we use the first 4 maps, which are the Body Center heatmaps and the Localization Offset maps in the front view and bird's-eye view. 
We first expand the front-/bird's-eye-view maps in depth/height  and then combine them to generate the 3D Center/Offset maps.
For coarse detection, we extract the rough 3D position of people from the 3D Center map. 
For fine localization, we sample the offset vectors from the 3D Offset map at the corresponding 3D center position. 
Adding these gives the 3D translation prediction.
For 3D mesh parameter regression, we use the estimated 3D translation $(x_i, y_i, d_i)$ and the Mesh Feature map.
The depth value $d_i$ of 3D translation is mapped to a depth encoding.
At $(x_i, y_i)$, we sample a feature vector from the Mesh Feature map and add it to the depth encoding for final parameter regression.
Finally, we convert the estimated parameters to body meshes using the SMPL+A model.

\subsection{SMPL+A: Mesh Representation for All Ages}

The SMPL~\cite{smpl} and SMIL~\cite{hesse2018learning} models are developed to parameterize 3D body meshes of adults and infants into low-dimensional parameters. Recently, AGORA~\cite{patel2021agora} further extends SMPL to support children by linearly blending the SMIL and SMPL template shapes with a weight $\alpha\in[0,1]$, which we refer to as an ``age offset."
While blending the templates to address scale and proportion differences between adults and children, AGORA uses the adult shape space regardless of age. 
Additionally, AGORA does not address the representation of infants.
We make a small, but important, change to better support all ages.

Following the notation of SMPL \cite{smpl}, the SMPL+A model defines a piece-wise function $\vec{B}=\mathcal{M}(\vec{\theta},\vec{\beta},\alpha)$ that maps 3D pose $ \vec{\theta}$, shape $ \vec{\beta}$, and age offset $\alpha$ to a 3D body mesh $\vec{B}\in\mathbb{R}^{6890 \times 3}$.
The pose parameters, $ \vec{\theta} \in \mathbb{R}^{6 \times 22}$, correspond to the 6D  rotations \cite{Zhou_2019_CVPR} of the first 22 body joints of SMPL.
The shape parameter $\vec{\beta} \in \mathbb{R}^{10}$ are the top-10 PCA coefficients of either the SMPL gender-neutral shape space or the SMIL shape space.

The adult shape space of AGORA produces shape deformations that are too large for an infant body, resulting in a distorted mesh when posed.
Therefore, we use SMIL for infants when the age offset $\alpha$ is above a threshold $t_{\alpha}$.
When $\alpha$ \textgreater $t_{\alpha}$, $\mathcal{M}(\vec{\theta},\vec{\beta},\alpha)$ is the SMIL model $\mathcal{M_I}(\vec{\theta},\vec{\beta})$.
When the age offset $\alpha \leq t_{\alpha}$, we use the AGORA formulation
\begin{equation}
\begin{aligned}
      \mathcal{M}(\vec{\theta},\vec{\beta},\alpha)=W(T_A(\vec{\theta},\vec{\beta},\alpha;\boldsymbol{\overline{T}},\boldsymbol{T_I}), J(\vec{\beta}), \vec{\theta}, \boldsymbol{\mathcal{W}}),\\
      T_A(\cdot)=
   (1-\alpha)\boldsymbol{\overline{T}}+\alpha\boldsymbol{\overline{T}_I}+B_S(\vec{\beta})+B_P(\vec{\theta}),
\end{aligned}
\label{eq:SMPL+A}
\end{equation}
where $W(\cdot)$ performs linear blend-skinning with weights $\boldsymbol{\mathcal{W}}$ to convert the T-posed mesh $T_A(\cdot)$ to the target pose $\vec{\theta}$ based on the skeleton joints $J(\cdot)$.
The T-posed mesh $T_A(\cdot)$ is the weighted sum of the templates ($\boldsymbol{\overline{T}},\boldsymbol{\overline{T}_I}$), shape-dependent deformation $B_S(\cdot)$, and pose-dependent deformation $B_P(\cdot)$. 
The age offset $\alpha \in [0,1]$ is used to interpolate between the adult SMPL template $\boldsymbol{\overline{T}}$ and the infant SMIL template $\boldsymbol{\overline{T}_I}$. 
The larger the $\alpha$, the lower the mesh template height.

The 3D joints $\vec{J}$ of the output mesh are derived via $\mathcal{J}\vec{B}$, where $\mathcal{J} \in \mathbb{R}^{K \times 6890}$ is a sparse weight matrix that linearly maps the vertices $\vec{B}$ to the $K$ body joints.
To supervise 3D joints $\vec{J}$ with 2D keypoints, regression methods~\cite{hmr,romp} typically adopt a weak-perspective camera model to project $\vec{J}$ into the image plane. 
For better depth reasoning, we employ a perspective camera model to perform projection; see Sup.~Mat.~for the details of our camera model.

\begin{figure}[t]
	\centerline{\includegraphics[width=0.98\columnwidth]{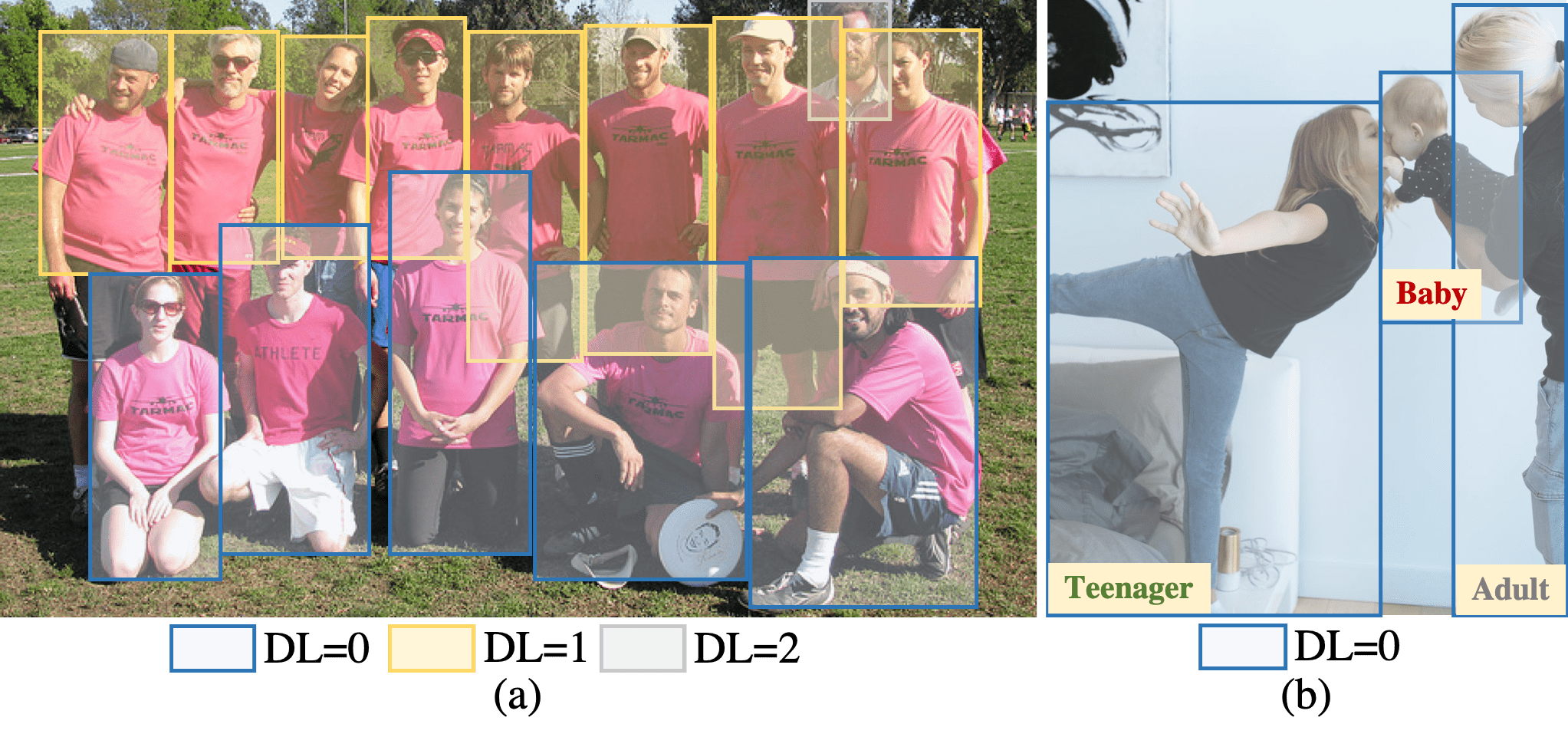}}
	\vspace{-5mm}
	\caption{Example images from the Relative Human (RH) dataset with weak annotations: depth layers (DLs) and age group classification. Examples are a) adults at different DLs, and b) people of different age groups at the same DL.}
	\label{fig:RH_dmeos}\vspace{-3mm}
\end{figure}

\subsection{Relative Human dataset}
\label{sec:relative_human}
Existing in-the-wild datasets lack groups of overlapping people with annotations.
Since acquiring 3D annotations of large crowds is challenging, we exploit more cost-effective weak annotations. 
We collect a new dataset, named Relative Human (RH), to support in-the-wild monocular human depth reasoning.

The images are collected from multiple sources to ensure diversity in age, ethnicity, gender, and scene.
Most images are collected from the existing 2D pose datasets~\cite{coco,crowdpose,zhang2019pose2seg}. 
They contain few infants so we collect additional open-source family photos from Pexels~\cite{pexels} and then annotate their 2D poses.
As shown in Fig.~\ref{fig:RH_dmeos}, we annotate the relative depth relationship between all people in the image.
We treat subjects whose depth difference is less than one body-width ($\gamma=0.3m$) as people in the same layer.
We then classify all people into different depth layers (DLs).
Unlike prior work, which labels the ordinal relationships between pairs of joints of individuals~\cite{chen2016single}, DLs capture the depth order of multiple people.
Additionally, we label people with four age categories: adults, teenagers, children, and babies.

In total, we collect about 7.6K images with weak annotations of over 24.8K people.
More than $21\%$ of the subjects are young people (5.3K), including teenagers, children, and babies.
For more analysis, please refer to Sup.~Mat.

\subsection{Representations}\label{sec:representations}
\noindent Figure~\ref{fig:framework} gives an overview of BEV's representations.

\textbf{Heatmaps}:
We build on the body-center heatmap representation from ROMP \cite{romp}.
The front-view heatmap of size $\mathbb{R}^{1 \times H \times W}$ is aligned with the pixel space and represents the likelihood of a body being centered at a 2D location using Gaussian kernels.
We go beyond ROMP to add a second 2D heatmap of size $\mathbb{R}^{1 \times D \times W}$ that represents an {\em unseen} bird's-eye-view.
This heatmap represents the likelihood of a person being at some point in depth; this map, however, does not represent metric depth.
BEV composes and refines these two maps into a 3D heatmap, $\boldsymbol{M_C^{3D}} \in  \mathbb{R}^{1\times D \times H \times W}$, which represents the 3D position of the detected human body centers with 3D Gaussian kernels.


\textbf{Offset maps}:
The discretized Center Heatmaps coarsely localize the body but we want the network to produce more precise estimates.
To improve the granularity of 3D localization, we use additional maps that, at each position, add an estimated offset vector to refine the coarse detection.
The front-view Offset map of size $\mathbb{R}^{3 \times H \times W}$ contains 3D offset vectors.
The bird's-eye-view Offset map of size $\mathbb{R}^{1 \times D \times W}$ contains 1D offset vectors for depth correction.
$\boldsymbol{M_O^{3D}} \in \mathbb{R}^{3 \times D \times H \times W}$ corresponds to the 3D Center map and contains a 3D offset vector at each 3D position.

\begin{figure}[t]
	\centerline{	\includegraphics[width=0.92\columnwidth]{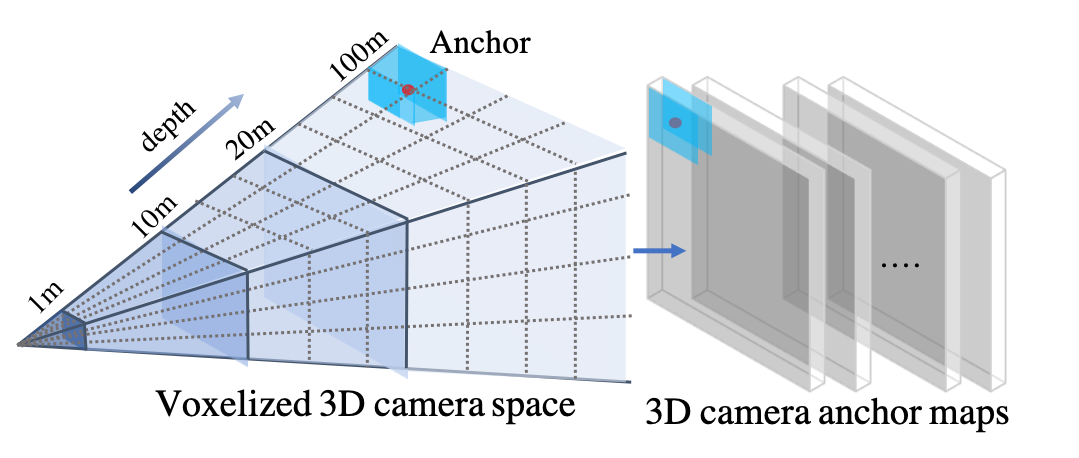}}
	\vspace{-0.1in}
	\caption{Pre-defined 3D camera anchor maps.}
	\label{fig:camera_anchors}\vspace{-3mm}
\end{figure}

\textbf{3D camera anchor maps}:
Each discretized coordinate in the 3D Center map corresponds to a set of camera parameters, representing its 3D position in the world.
The anchor map serves as a mapping function to transform the coordinates of the 3D Center map to the 3D position in a predefined perspective camera space.
To establish a one-to-one mapping from the square Center map to a pyramidal camera space, as shown in Fig.~\ref{fig:camera_anchors}, we voxelize camera space.
Each voxel center corresponds to a discretized 3D coordinate in the Center map.
The 3D position vector $(x,y,d)$ of voxel center is the anchor value of 3D camera anchor map.
Voxels of equal depth form a depth plane, corresponding to a 2D (x-y) slice of the 3D camera anchor map.
During inference, the 3D camera anchor map is sampled at the same coordinate of 3D Center map to obtain the coarse 3D translation of the corresponding detection. 

\textbf{Mesh feature map}:
$\boldsymbol{M_F} \in \mathbb{R}^{128 \times H \times W}$ contains a 128-D mesh feature vector at each 2D position.
These features are aligned with the input 2D image at the pixel level. 
After a 3D-center-based sampling process, the relevant features are used for the regression of SMPL+A parameters.

\subsection{BEV}
To effectively establish the 3D representation, the front-view and the bird's-eye-view must work together to estimate the image position and depth of corresponding subjects.
Independently estimating the map of two views in parallel would inevitably cause misalignment, leading to the failure of 3D heatmap-based detection. 
To connect the two views, we estimate the bird's-eye-view maps conditioned on the front-view maps (i.e.~Center and Offset maps).
Specifically, to estimate the bird's-eye-view maps, we take the concatenation of the front-view maps and the backbone feature maps as input.
The front-view 2D body-centered heatmap is used as a form of robust attention to people in the image, which helps the model focus on exploring depth during bird's-eye view estimation.
Then we expand and composite the 2D maps from the front and BEV views to generate the 3D maps.
To integrate 2D features from two views and enhance 3D consistency, we further perform 3D convolution on the composited 3D maps for refinement.

Next, we extract the 3D translation from the estimated 3D maps, $\boldsymbol{M_C^{3D}},\boldsymbol{M_O^{3D}}$.
High-confidence 3D positions of the 3D Center map are where we sample 3D offset vectors from the 3D Offset map. 
From the same 3D position in the 3D camera anchor maps (Fig.~\ref{fig:camera_anchors}), we obtain the 3D anchor values, which are positions in camera space of the corresponding 3D center voxel. 
Adding the 3D offset vectors to the 3D anchor values gives the 3D translation as output. 

Finally, we take the estimated 3D translation $(x_i, y_i, d_i)$ and Mesh Feature maps $\boldsymbol{M_F}$ for parameter regression.
We sample the pixel-level mesh feature vectors at $(x_i, y_i)$ of $\boldsymbol{M_F}$. 
Inspired by positional embeddings~\cite{transformer}, we learn an embedding space to differentiate people at different depths, especially for the overlapping cases.
The predicted depth value $d_i$ is mapped to a 128-dim encoding vector via an embedding layer.
We sum up the depth encodings and the mesh feature vectors to differentiate the features of people at different depths, enabling individual estimates for different subjects.
Then we estimate the SMPL+A parameters $(\vec{\theta},\vec{\beta},\alpha)$ via a fully-connected block.
The output body meshes are obtained via $\mathcal{M}(\vec{\theta},\vec{\beta},\alpha)$. 

\subsection{Loss Functions}\label{sec:loss_functions}

Our loss functions are divided into two groups illustrated in Fig.~\ref{fig:framework}: relative losses (in gold) and the standard mesh losses (in black).
BEV is supervised by the weighted sum of all loss items.
First, we introduce two relative loss functions for weakly supervised training (WST).

\textbf{Piece-wise depth layer loss $\boldsymbol{\mathcal{L}_{depth}}$.}
$\boldsymbol{\mathcal{L}_{depth}}$ is designed to supervise the predicted depth $d_i, d_j$ of subject $i, j$ by their depth layers $r_i, r_j$ via
\begin{equation}
\setlength{\abovedisplayskip}{1pt}
\setlength{\belowdisplayskip}{1pt}
\small
\begin{aligned}
\begin{cases} 
      (d_i-d_j)^2, & r_i=r_j\\
      log(1+e^{d_i-d_j})\prod((d_i-d_j)-\gamma(r_i-r_j)), & r_i < r_j \\
      log(1+e^{d_j-d_i})\prod(\gamma(r_i-r_j)-(d_i-d_j)), & r_i > r_j ,\\
\end{cases}
\end{aligned}
\label{eq:hierarchical_depth}
\end{equation}
where $\prod$ is a binarization function that maps positive values to 1 and negative values to 0.
$\prod$ is used to judge whether the BEV prediction is consistent with the depth relationship of the ground truth  DLs.
$\boldsymbol{\mathcal{L}_{depth}}$ is 0, if the predicted depth difference is within an acceptable range; that is, greater than the product of the DL difference and body-width $\gamma$.
Otherwise, $\boldsymbol{\mathcal{L}_{depth}}$ will encourage the model to achieve it.

Previous ordinal depth losses~\cite{chen2016single,pavlakos2018ordinal} encourage the model to enlarge the depth difference between people at different depth layers as much as possible.
In contrast, the penalty in $\boldsymbol{\mathcal{L}_{depth}}$ is controlled within a  range.
This helps avoid pushing remote subjects too far away.

\textbf{Ambiguity-compatible age loss $\boldsymbol{\mathcal{L}_{age}}$.}
The classification of age categories (infant, child, teenager, adult) is inherently ambiguous, especially for teenagers and children.
Also, while height is correlated with age, one can easily find children who are taller than some adults.
Consequently, we formulate an ambiguity-compatible mixed loss $\boldsymbol{\mathcal{L}_{age}}$. 

Rather than supervise height directly, we supervise the $\alpha$ parameter that controls the blending between the SMIL infant body and the SMPL adult body.
To do so, we define ranges of $\alpha$ values for each age group; i.e.~(lower-bound, middle, upper-bound).
We do this using the statistical data of heights for each age category that we then relate these to ranges of $\alpha$ values.
Formally, the ranges are  $(\alpha_{l}^k,\alpha_{m}^k,\alpha_{u}^k),k=1\cdots4$ where $k$ is the annotated age class number; see Sec.~\ref{sec:experiments} for details.

BEV is then trained to predict the body shape as well as an $\alpha$ value for each person.
Given the predicted $\alpha$ and ground truth age class $k_{g}$, the loss $\boldsymbol{\mathcal{L}_{age}}$ is defined as
\begin{equation}
\setlength{\abovedisplayskip}{1pt}
\setlength{\belowdisplayskip}{1pt}
\small
\begin{aligned}
\boldsymbol{\mathcal{L}_{age}}(\alpha)=
\begin{cases} 
      0, & \alpha_{l}^{k_{g}}< \alpha \leq \alpha_{u}^{k_{g}}\\
      (\alpha-\alpha_{m}^{k_{g}})^2, & \mathrm{otherwise.}
\end{cases}
\end{aligned}
\label{eq:age_loss}
\end{equation}

\textbf{Other losses.}
Following the previous methods~\cite{romp,hmr}, we employ the standard mesh losses to supervise the output maps and regressed SMPL+A parameters. 
$\boldsymbol{\mathcal{L}_{cm}}$ is the focal loss~\cite{romp} of the front-view Body Center heatmap.
In the same pattern, we further use a 3D focal loss $\boldsymbol{\mathcal{L}_{cm3D}}$ to supervise the 3D Center map via converting $\boldsymbol{\mathcal{L}_{cm}}$'s 2D operation to 3D. 
$\boldsymbol{\mathcal{L}_{pm}}$ consists of three parts, $\boldsymbol{\mathcal{L}_{\theta}},\boldsymbol{\mathcal{L}_{\beta}}$, and $\boldsymbol{\mathcal{L}_{prior}}$.
$\boldsymbol{\mathcal{L}_{\theta}}$ and $\boldsymbol{\mathcal{L}_{\beta}}$ are $L_2$ losses of SMPL+A pose $\vec{\theta}$ and shape $\vec{\beta}$ parameters respectively.
$\boldsymbol{\mathcal{L}_{prior}}$ is the Mixture of Gaussian pose prior~\cite{keep,smpl} on $\vec{\theta}$.
To supervise the 3D body joints $\vec{J}$, we use $\boldsymbol{\mathcal{L}_{j3d}}$, which is composed of $\boldsymbol{\mathcal{L}_{mpj}}$ and $\boldsymbol{\mathcal{L}_{pmpj}}$.
$\boldsymbol{\mathcal{L}_{mpj}}$ is the $L_2$ loss of 3D joints $\vec{J}$.
To alleviate the domain gap between training datasets, we follow~\cite{romp,sun2019dsd-satn} to calculate the $L_2$ loss $\boldsymbol{\mathcal{L}_{pmpj}}$ of the predicted 3D joints after Procrustes alignment with the ground truth.
$\boldsymbol{\mathcal{L}_{pj2d}}$ is the $L_2$ loss of the 2D projection of 3D joints $\vec{J}$.
Lastly, $w_{(.)}$ denotes the corresponding weight of these losses.

\begin{table*}[t]
\setlength\tabcolsep{1.5mm}
\hspace{2mm}
\parbox{.5\linewidth}{
\centering
	\footnotesize
	\begin{tabular}{l|ccccc|c}
	\toprule
	\multirow{2}{*}{Method} & \multicolumn{5}{c|}{{PCDR$^{0.2}$}(\%)$\uparrow$} & \multirow{2}{*}{{mPCK$_{h}^{0.6}$}$\uparrow$} \\
	\cline{2-6}
	& \multicolumn{1}{c}{Baby} & \multicolumn{1}{c}{Kid} & \multicolumn{1}{c}{Teen} & \multicolumn{1}{c}{Adult} & \multicolumn{1}{c|}{All} & \\
    \midrule
    3DMPPE$^\dagger$~\cite{moon2019camera} & 39.33 & 51.42 & 60.91 & 57.95 & 57.47 & - \\ %
    CRMH~\cite{jiang2020coherent} & 34.74 & 48.37 & 59.11 & 55.47 & 54.83 & 0.781  \\ 
    SMAP~\cite{zhen2020smap} & 31.58 & 40.29 & 47.35 & 41.65 & 41.55 & - \\ 
    ROMP~\cite{romp}& 30.08 & 48.41 & 51.12 & 55.34 & 54.81 & 0.866 \\ 
    BEV w/o WST & 34.27 & 50.81 & 54.34 & 57.43 & 57.17 & 0.850 \\ 
    BEV w/o $\boldsymbol{\mathcal{L}_{depth}}$ & 43.61 & 51.55 & 50.88 & 57.27 & 55.97 & 0.794 \\ 
    BEV w/o $\boldsymbol{\mathcal{L}_{age}}$ & 49.09 & 56.55 & 60.92 & 62.47 & 61.47 & 0.810 \\ 
    BEV & \textbf{60.77} & \textbf{67.09} & \textbf{66.07} & \textbf{69.71} & \textbf{68.27} & \textbf{0.884} \\ 
	\bottomrule
    \end{tabular}
    \vspace{-2mm}
	\caption{Accuracy of relative depth relations (PCDR$^{0.2}$) and projected 2D poses (mPCK$_h^{0.6}$) on RH. $^\dagger$ uses the ground truth bounding boxes. }
	\label{tab:relative_human}}
\hspace{5mm}
\parbox{.4\linewidth}{
\centering
\footnotesize
    \begin{tabular}{l|cccc|c}
    \toprule
    Method & Haggl. & Mafia & Ultim. & \multicolumn{1}{c|}{Pizza} & Mean\\
   \midrule
        Zanfir et. al.~\cite{zanfir2018deep} & 141.4 & 152.3 & 145.0 & 162.5 & 150.3 \\
        MSC~\cite{zanfir2018monocular} & 140.0 & 165.9 & 150.7 & 156.0 & 153.4 \\
        CRMH~\cite{jiang2020coherent} & 129.6 & 133.5 & 153.0 & 156.7 & 143.2 \\
        ROMP~\cite{romp} & 110.8 & 122.8 & 141.6 & 137.6 & 128.2\\
        3DCrowdNet~\cite{choi20223dcrowdnet} & 109.6 & 135.9 & 129.8 & 135.6 & 127.3 \\
        \textbf{BEV} & \textbf{90.7} & \textbf{103.7} & \textbf{113.1} & \textbf{125.2} & \textbf{109.5}\\
   \bottomrule
    \end{tabular}
    \vspace{-3mm}
    \caption{Comparisons to the state-of-the-art methods on CMU Panoptic in MPJPE. Results are obtained from the original papers.}  \label{tab:CMU Panoptic}}
\end{table*}\vspace{-3mm}

\begin{table*}[t]
\setlength\tabcolsep{3pt}
	\footnotesize
	\centering
	\begin{tabular}{l|ccc|cc|cc|ccc|cc|cc}
	\toprule
 \multirow{3}{*}{Method}&  \multicolumn{7}{c|}{Kid subset} & \multicolumn{7}{c}{Full set} \\
	\cline{2-15}
& \multicolumn{3}{c|}{Detection$\uparrow$} & \multicolumn{2}{c|}{Matched$\downarrow$}  &  \multicolumn{2}{c|}{All$\downarrow$} & \multicolumn{3}{c|}{Detection$\uparrow$} & \multicolumn{2}{c|}{Matched$\downarrow$}  &  \multicolumn{2}{c}{All$\downarrow$} \\
\cline{2-15}
&  \multicolumn{1}{c}{F1 score} & \multicolumn{1}{c}{Precision} & \multicolumn{1}{c|}{Recall} & \multicolumn{1}{c}{MVE} & \multicolumn{1}{c|}{MPJPE} & \multicolumn{1}{c}{NMVE} & \multicolumn{1}{c|}{NMJE} &  \multicolumn{1}{c}{F1 score} & \multicolumn{1}{c}{Precision} & \multicolumn{1}{c|}{Recall} & \multicolumn{1}{c}{MVE} & \multicolumn{1}{c|}{MPJPE} & \multicolumn{1}{c}{NMVE} & \multicolumn{1}{c}{NMJE} \\
\midrule
PARE~\cite{kocabas2021pare}& 0.55 & 0.44 & 0.74 & 186.4 & 193.9 & 338.9 & 352.5 & 0.84 & 0.96 & 0.75 & 140.9 & 146.2 & 167.7 & 174.0\\
SPIN~\cite{patel2021agora}& 0.31 & 0.21 & 0.60 & 186.7 & 191.7 & 602.3 & 618.4 & 0.77 & 0.91 & 0.67 & 148.9 & 153.4 & 193.4 & 199.2\\
SPEC~\cite{Kocabas_SPEC_2021} & 0.52 & 0.40 & 0.73 & 163.2 & 171.0 & 313.8 & 328.8 & 0.84 & 0.96 & 0.74 & 106.5 & 112.3 & 126.8 & 133.7 \\
ROMP~\cite{romp} & 0.50 & 0.37 & 0.80 & 156.6 & 159.8 & 313.2& 319.6 & 0.91 & 0.95 & 0.88 & 103.4 & 108.1 & 113.6 & 118.8\\
BEV w/o WST  & \textbf{0.58} & \textbf{0.44} & \textbf{0.86} & 146.0 & 148.3 & 251.7 & 255.7 & 0.93 & 0.96 & 0.90 & 105.6 & 109.7 & 113.5 & 118.0\\
BEV  & 0.55 & 0.41 & 0.85 & \textbf{125.9} &\textbf{129.1} & \textbf{228.9} & \textbf{234.7} & \textbf{0.93} & \textbf{0.96} & \textbf{0.90} & \textbf{100.7} & \textbf{105.3} & \textbf{108.3} & \textbf{113.2} \\
\bottomrule
\end{tabular}
\vspace{-0.1in}
	\caption{Comparison of SOTA methods on AGORA test set. All methods are fine-tuned on the AGORA training set or synthetic data~\cite{Kocabas_SPEC_2021} generated in the same way as AGORA. We fine-tune ROMP~\cite{romp} using the public implementation; results from  the AGORA leaderboard.}
	\label{tab:agora}
\end{table*}

\section{Experiments}
\label{sec:experiments}

\subsection{Implementation Details}

\textbf{Training details.}
For basic training, we use two 3D pose datasets (Human3.6M~\cite{h36m} and MuCo-3DHP~\cite{mehta2018single}) and four 2D pose datasets (COCO~\cite{coco}, MPII~\cite{mpii}, LSP~\cite{lsp_extended}, and CrowdPose~\cite{crowdpose}). 
We also use the pseudo SMPL annotations from~\cite{joo2020eft} and WST on RH.
Most samples in RH are collected from 2D pose datasets~\cite{coco,crowdpose,zhang2019pose2seg}.
For a fair comparison, we only use the samples that are also used for training in compared methods~\cite{romp,moon2019camera,jiang2020coherent,zhen2020smap,kolotouros2019spin,Kocabas_SPEC_2021}.
To compare with~\cite{Kocabas_SPEC_2021,patel2021agora}, we further fine-tune our model and ROMP on AGORA.
The threshold for the age offset is set to $t_{\alpha}=0.8$.
The age offset ranges $(\alpha_{l}^k,\alpha_{m}^k,\alpha_{u}^k)$ are: adults $(-0.05, 0, 0.15)$, teenagers $(0.15, 0.3, 0.45)$, children $(0.45, 0.6, 0.75)$, and infants $(0.75, 0.9, 1)$.
See Sup.~Mat.~for more details. 

\textbf{Evaluation benchmarks.}
We evaluate BEV on three multi-person datasets, RH, CMU Panoptic, \cite{cmu_panoptic} and AGORA~\cite{patel2021agora}, containing 257 child scans and significant person-person occlusion. 

\textbf{Evaluation matrix.}
To evaluate the accuracy of depth reasoning, we employ the Percentage of Correct Depth Relations (\textbf{PCDR$^{0.2}$}), and set the threshold for equal depth to $0.2m$.
To evaluate the accuracy of projected 2D poses on RH, we also report the mean Percentage of Correct Keypoints (\textbf{mPCK$_{h}^{0.6}$}),  setting the matching threshold to $0.6$ times the head length.

Also, following AGORA~\cite{patel2021agora}, we evaluate the accuracy of 3D pose/mesh estimation while considering missing detections.
To evaluate the detection accuracy, we report \textbf{Precision}, \textbf{Recall}, and \textbf{F1 score}. 
For matched detections, we report the Mean Per Joint Position Error (\textbf{MPJPE}) and Mean Vertex Error (\textbf{MVE}). 
To punish misses and false alarms in detection, we normalize the MPJPE and MVE by F1 score to get Normalized Mean Joint Error (\textbf{NMJE}) and Normalized Mean Vertex Error (\textbf{NMVE}).

\subsection{Comparisons to the state-of-the-art methods}

\textbf{Monocular depth reasoning.}
We first evaluate BEV on monocular depth reasoning in Tab.~\ref{tab:relative_human} using the RH dataset.
Results in Tab.~\ref{tab:relative_human} are obtained using the official implementations of compared methods. 
BEV uses the same training samples as~\cite{romp} to perform WST.
We first compare with the most competitive methods~\cite{moon2019camera,jiang2020coherent,zhen2020smap}, which solve depth relations in monocular images.
We also compare with ROMP~\cite{romp}, for one-stage multi-person mesh recovery.
Their 3D translation results are obtained by solving the PnP algorithm (RANSAC~\cite{fischler1981random}) between their 3D pose and projected 2D pose predictions.
As shown in Tab.~\ref{tab:relative_human}, BEV outperforms all these methods in the accuracy of both depth reasoning and projected 2D poses by a large margin.

\begin{figure*}[t]
	\centerline{	\includegraphics[width=1\textwidth]{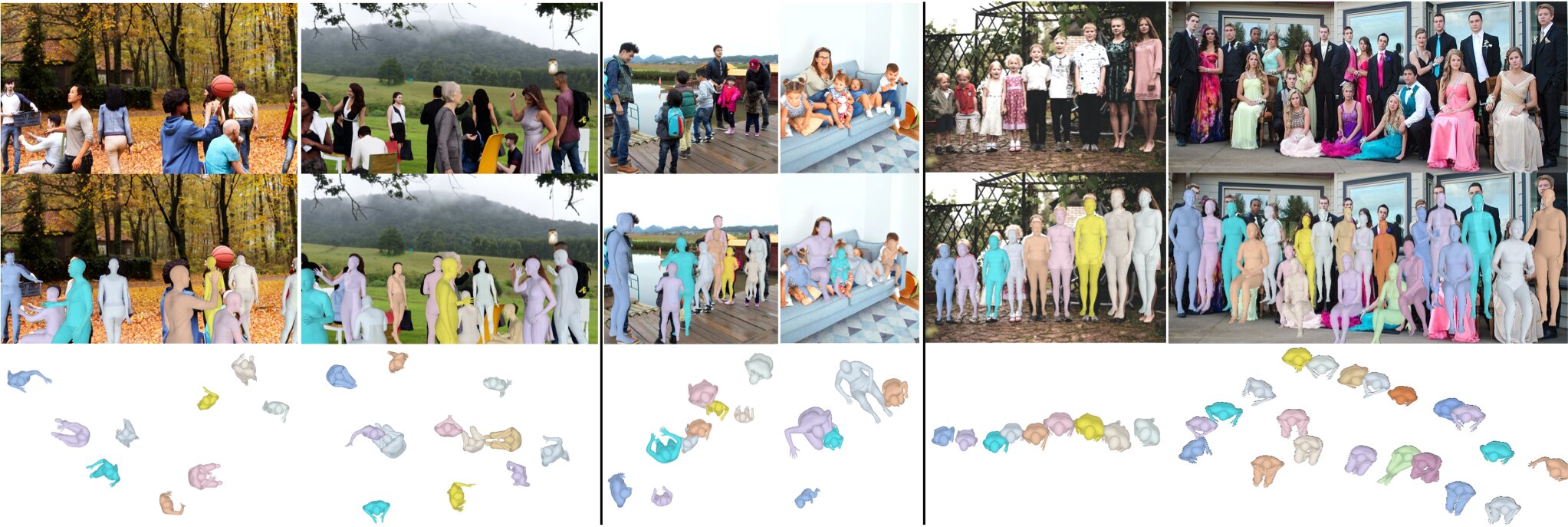}}
	\vspace{-0.1in}
	\caption{Qualitative results on AGORA, RH, and Internet images~\cite{pexels}. Note how children and adults are properly placed in depth.}\vspace{-1mm}
	\label{fig:qualitative_results}
\end{figure*}

\textbf{Monocular detection and mesh regression.}
We also run BEV on AGORA and CMU Panpotic to evaluate the detection and 3D mesh accuracy.
We compare with the state-of-the-art (SOTA) multi-stage methods~\cite{zanfir2018deep,zanfir2018monocular,jiang2020coherent,kocabas2021pare,Kocabas_SPEC_2021,patel2021agora,choi20223dcrowdnet} and the one-stage ROMP~\cite{romp}.
Benefiting from the superiority in recall, in Tab.~\ref{tab:agora}, BEV outperforms SOTA methods on detection by $5.2\%$ and $2.2\%$ in terms of F1 score on the kid and full subset, respectively.
This is evidence that the 3D representation helps alleviate depth ambiguity in crowded scenes.
On the kid subset, BEV significantly outperforms previous methods in terms of mesh reconstruction.
Especially, compared with ROMP~\cite{romp}, BEV reduces errors over $19.6\%$ and $26.9\%$ in terms of matched MVE and all NMVE on AGORA kids, indicating that BEV effectively reduces the age bias using WST.
Also, as shown in Tab.~\ref{tab:CMU Panoptic}, on CMU Panpotic, BEV significantly reduces 3D pose errors by $13.9\%$ compared to multi-person SOTA methods.
For qualitative results, see Fig.~\ref{fig:teaser} and
Fig.~\ref{fig:qualitative_results}.

\subsection{Ablation Studies}



\textbf{Bird's-eye-view representation \&\/ BEV w/o WST.}
To further test the effectiveness of BEV's 3D representation, we train it without performing WST on RH and compare it with SOTA methods on AGORA and RH.
On RH in Tab.~\ref{tab:relative_human}, compared with CRMH~\cite{jiang2020coherent}, the depth reasoning accuracy of BEV w/o WST is 4.1\% higher (\textbf{PCDR$^{0.2}$} of all).
BEV w/o WST outperforms the 2D representation-based network ROMP~\cite{romp}.
These results point to the effectiveness of our 3D representation for dealing with monocular depth ambiguity.
On AGORA, as shown in Tab.~\ref{tab:agora}, BEV w/o WST significantly outperforms ROMP in all detection metrics.
Additionally, the strong detection ability of the 3D representation makes BEV w/o WST outperform the SOTA methods~\cite{romp,Kocabas_SPEC_2021,patel2021agora} in terms of NMVE and NMJE.

\textbf{Weakly supervised training (WST) losses, $\boldsymbol{\mathcal{L}_{depth}}$ and $\boldsymbol{\mathcal{L}_{age}}$.}
Results in Tab.~\ref{tab:relative_human} show that performing WST significantly improves the accuracy of depth reasoning, especially for the young groups.
Also, Tab.~\ref{tab:relative_human} shows that separately using $\boldsymbol{\mathcal{L}_{depth}}$ or $\boldsymbol{\mathcal{L}_{age}}$ make BEV produce better depth reasoning than BEV w/o WST, and, when using both terms, BEV performs best.

\textbf{3D Offset map (OM) and Front-view condition (FVC) for 3D localization}. 
FVC is taking the front-view 2D body-centered heatmap as a robust attention signal to explore the depth of detected persons during bird's-eye view estimation. 
Results in Tab.~\ref{tab:ablation_study} verify that OM and FVC significantly improve the granularity of 3D localization.

\textbf{Piece-wise depth layer loss $\boldsymbol{\mathcal{L}_{depth}}$ v.s. ordinal depth loss~\cite{wang2020hmor}.}
Unlike an ordinal depth loss, $\boldsymbol{\mathcal{L}_{depth}}$ keeps the penalty within a reasonable range (see Sec.~\ref{sec:loss_functions}).
As shown in Tab.~\ref{tab:Loss}, on AGORA validation set, training with $\boldsymbol{\mathcal{L}_{depth}}$ reduces the 3D translation error, especially in depth.

\begin{table}[t]
\footnotesize
\centering
	\begin{tabular}{l|cc|ccc}
	\toprule
	\multirow{2}{*}{ Method}  &  \multicolumn{2}{c|}{ Relative Human} & \multicolumn{3}{c}{ AGORA} \\
	\cline{2-6}
	& \multicolumn{1}{c}{PCDR$^{0.2}$} & \multicolumn{1}{c|}{mPCK$_{h}^{0.6}$} & \multicolumn{1}{c}{ F1} & \multicolumn{1}{c}{ NMVE} & \multicolumn{1}{c}{ NMJE} \\
    \midrule
    BEV & \textbf{68.27} & \textbf{0.884} & \textbf{0.93} & \textbf{108.3} & \textbf{113.2}\\ 
    { w/o FVC} & 67.99 & 0.880 & 0.89 & 118.9 & 123.0 \\
    { w/o OM} & 60.76 & 0.620 & 0.87 & 126.6 & 130.7 \\
	\bottomrule
    \end{tabular}
    \vspace{-2mm}
    \caption{Ablation study of front-view condition (FVC) and 3D Offset map (OM) on RH and AGORA.}\vspace{-2mm}
    \label{tab:ablation_study}
\end{table}

\begin{table}[t]
	\footnotesize
	\centering
	\begin{tabular}{l|c|ccc}
	\toprule
Method & Dist.$\downarrow$ & X$\downarrow$ & Y$\downarrow$& Depth$\downarrow$\\
\midrule
Ordinal loss~\cite{wang2020hmor} & 0.608 & 0.153 & 0.184 & 0.509 \\
Piece-wise $\boldsymbol{\mathcal{L}_{depth}}$ (\textbf{ours})  & \textbf{0.518} & \textbf{0.128} & \textbf{0.166} & \textbf{0.423} \\
\bottomrule
\end{tabular}
\vspace{-2mm}
	\caption{3D translation error on AGORA validation set with different depth losses.}\vspace{-2mm}
	\label{tab:Loss}
\end{table}

\vspace{-2mm}
\section{Conclusion, Limitations, Ethics, Risks}\vspace{-1mm}

In this paper, we introduce BEV, a unified one-stage method for monocular regression and depth reasoning of multiple 3D people.
By introducing a novel bird's eye view representation, we enable powerful 3D reasoning that reduces the monocular depth ambiguity.
Exploiting the correlation between body height and depth, BEV learns depth reasoning from complex in-the-wild scenes by exploiting relative depth relations and age group classification.
We make available an in-the-wild dataset to promote the training and evaluation of monocular depth reasoning in the wild.
The ablation studies point to the value of the 3D representation and the fine-grained localization in the network, the importance of our training scheme, and the value of the collected dataset. 
BEV is a preliminary attempt to explore complex multi-person relationships in the 3D world, and we hope the framework will serve as a simple yet effective foundation for future progress. 

\textbf{Limitations.} 
While BEV goes beyond current methods to cover more diverse ages, it is not trained to capture diverse weights, gender, ethnicity, etc. 
BEV also assumes a constant focal length. 
Our labeling approach, however, suggests that weak labels can produce strong results; i.e.~improved metric accuracy.
Note that BEV is not trained or designed to deal with large ``crowds" (e.g.~100's of people). 

\textbf{Ethics and data.}
We collected RH images from a free photo website \cite{pexels} under a Creative Commons license that enables sharing.
We strove to have a dataset that is diverse in age, ethnicity, and gender.
Also, our weak annotations do not contain any personal information and the annotators, themselves, are anonymous and were not studied.

\textbf{Potential Negative Societal Impacts.}
Methods for monocular 3D pose and shape estimation might be used for automated surveillance, tracking, and behavior analysis, which may violate people's privacy. 
To help prevent this, BEV is released for research only.

\noindent \textbf{Acknowledgements:} This work was supported by the National Key R\&D Program of China under Grand No. 2020AAA0103800. 

\noindent \textbf{Disclosure:} MJB has received research funds
from Adobe, Intel, Nvidia, Facebook, and Amazon and has financial interests in Amazon, Datagen Technologies, and Meshcapade GmbH. While he was part-time at Amazon during this project, his research was performed solely at Max Planck.

{\small
\bibliographystyle{ieee_fullname}
\bibliography{arxiv}
}

\end{document}


\title{Putting People in their Place: Monocular Regression of 3D People in Depth\\
\textbf{**Supplementary Material**}}


\twocolumn[{%
\renewcommand\twocolumn[1][]{#1}%
\maketitle\vspace{-2cm}
\begin{center}
    \centering
    \captionsetup{type=figure}
    \includegraphics[width=1\textwidth]{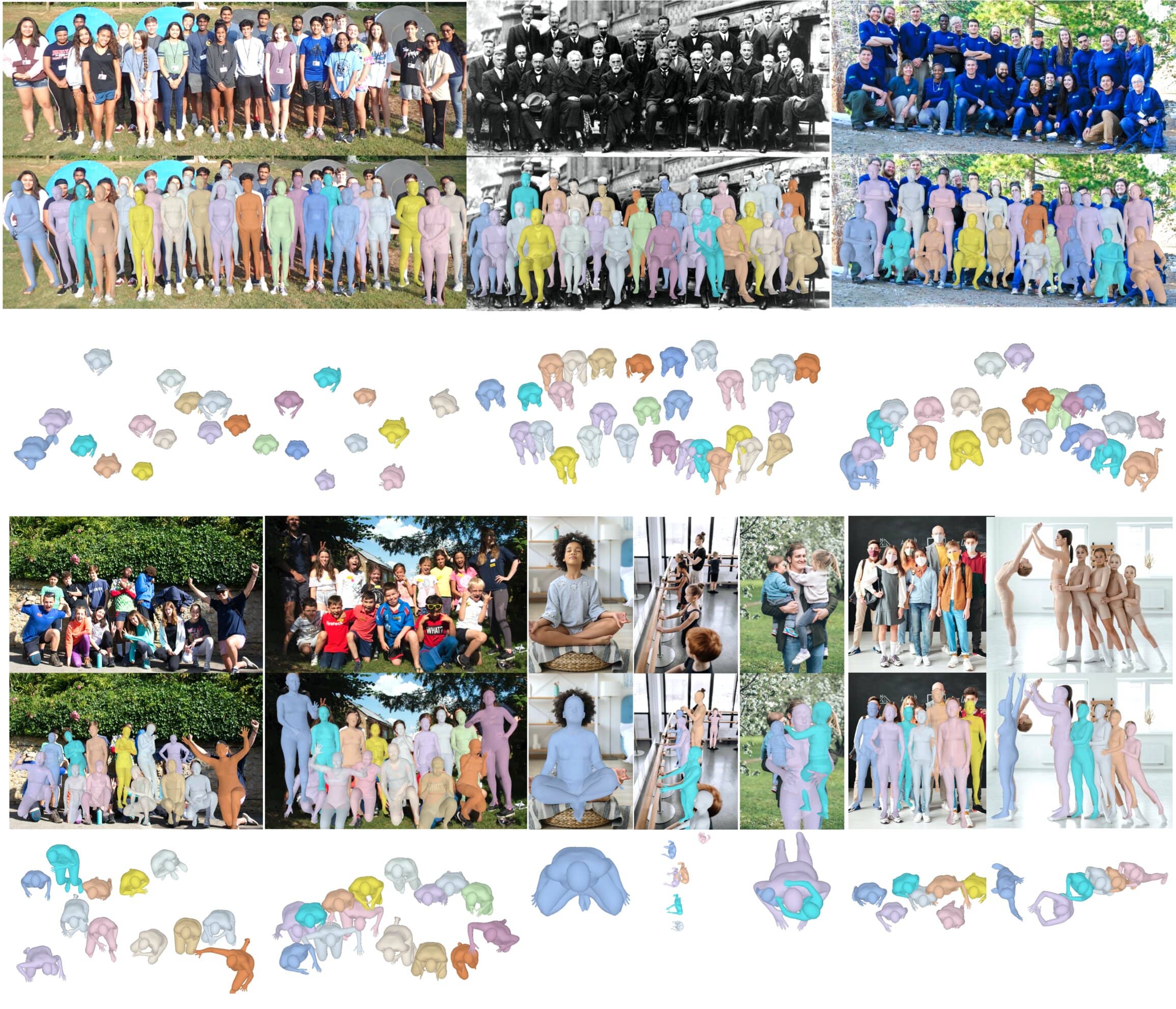}
    \captionof{figure}{More qualitative results on Internet images~\cite{pexels}.}
    \label{fig:internet}
\end{center}%
}]

\section{Introduction}
In this material, we provide more implementation details, analysis of the ``\textit{Relative Human}" (RH) dataset, and quantitative/qualitative comparisons to the state-of-the-art methods. 
Additionally, we present more visual results, like Fig.~\ref{fig:internet}, to show the performance of BEV under different situations and to explore its failure modes.

\section{Implementation Details}

In this section, we introduce the details of our camera representation, network architecture, and training details.

\subsection{Normalized Camera Representation~\label{sec:normalized_camera}}

To supervise 3D joints $\vec{J}$ with 2D poses, existing methods~\cite{hmr,romp} widely adopt a weak-perspective camera model to project $\vec{J}$ onto the image plane. 
For better depth reasoning, we employ a perspective camera model to perform this 2D projection. 

In most cases, accurate camera parameters for in-the-wild images are unavailable.
In this situation, to avoid reliance on the camera parameters of 2D projection, we assume that the input image is captured with a standard camera without radial distortion.
Then we can assign static values for the field of view (FOV) and image size $\vec{W}$ of this standard camera.
The focal length $\vec{f}=(f_x,f_y)$ can be defined as $\vec{W}/(2tan(FOV/2))$.
Given the 3D translation $(x_i,y_i,d_i)$ of $i$-th subject and the focal length, the 2D projection $(\vec{u_i},\vec{v_i})$ of 3D joints $(\vec{J_i^x},\vec{J_i^y},\vec{J_i^d})$ is defined as
\begin{equation}
\begin{aligned}
   \vec{u_i}=\frac{f_x(\vec{J_i^x}+x_i)}{\vec{J_i^d}+d_i}, \vec{v_i}=\frac{f_y(\vec{J_i^y}+y_i)}{\vec{J_i^d}+d_i} .
\end{aligned}
\label{eq:projection}
\end{equation}

In cases where the camera parameters are provided, we can convert the 3D translation estimated in our standard camera space to the given one.
With $K$ pairs of estimated 3D joints $\vec{J}$ and their 2D projection (obtained via Eq.~\ref{eq:projection}), we can solve the 3D translation at a specific camera space via a PnP algorithm (e.g. RANSAC~\cite{fischler1981random}).

However, in the image, 3D translation is not as intuitive as the person's scale used by weak-perspective methods.
For instance, a small 2D scale change in an image may correspond to a large difference in 3D translation in camera space, especially for  people who are far away in depth.
Therefore, to alleviate this difference, we convert the 3D translation $(x_i,y_i,d_i)$ to a normalized scale-based format $(s_i,t_i^y,t_i^x)$ via a scale factor $s_i=(d_i tan(FOV/2))^{-1}$, where $t_i^y=y_i s_i, t_i^x=x_i s_i$. 
The normalized representation is proportional to the person's scale. 
When FOV=60$^\circ$, the sensitive part $s_i \in (0,2)$ corresponds to $d_i \in (0.86,+\infty)$ in meters, which is more suitable for the network to estimate.

\begin{figure}[t]
	\centerline{\includegraphics[width=0.92\columnwidth]{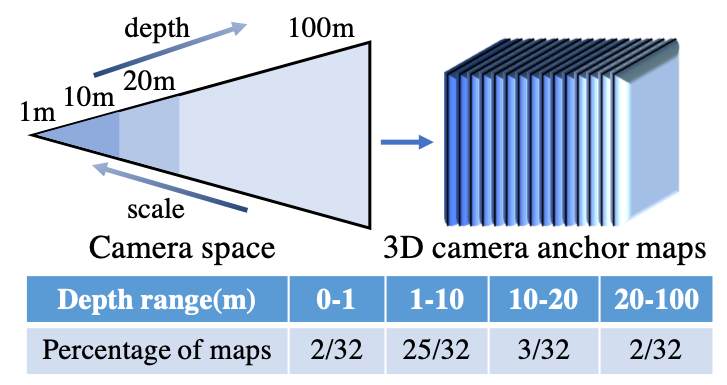}}
	\caption{Pre-defined 3D camera anchor maps.}
	\label{fig:camera_space_partition}
\end{figure}

Additionally, we observe that people in the depth range (1m,10m) show more abundant and stable information in pose, shape, and depth, which deserve more attention.
Additionally, most of our training samples are within this depth range.
As we introduced in the main paper, 3D camera anchor maps define the way we voxelize the 3D camera space. 
Therefore, we adjust the occupancy ratio of different depths in the channel number of 3D camera anchor maps.
As shown in Fig.~\ref{fig:camera_space_partition}, we first split the camera space into 4 regions in depth and then evenly put the different number (shown in the table) of 3D camera anchor maps inside each region.
For instance, we put 25/32 3D camera anchor maps inside the depth range (1m,10m); this gives more attention to this critical depth range. 
Each anchor map contains the normalized camera values $(s_i,t_i^y,t_i^x)$ at the corresponding position.

\begin{figure*}[t]
	\centerline{\includegraphics[width=1.00\textwidth]{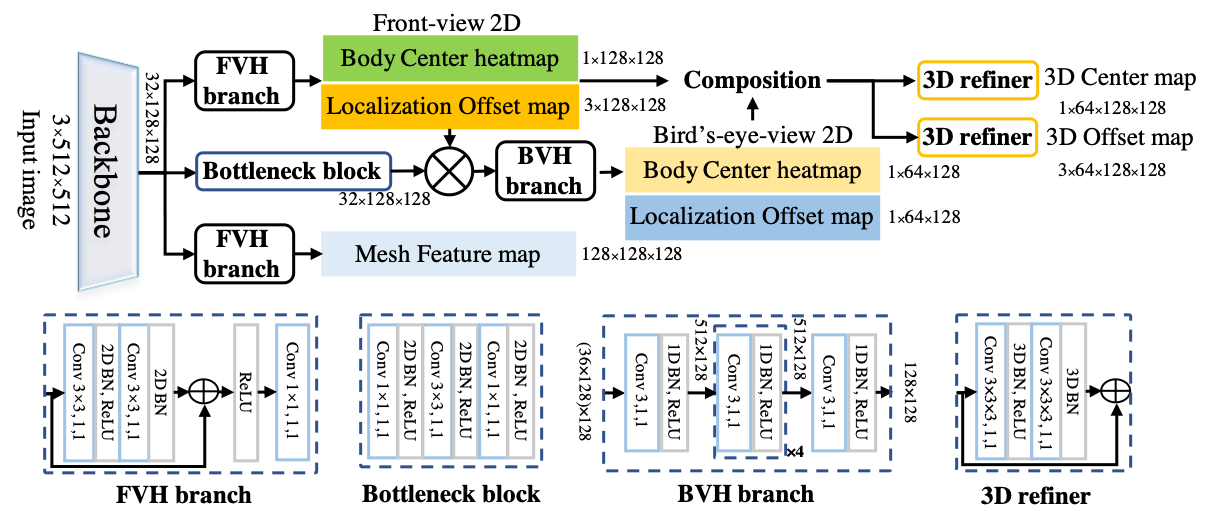}}
	\caption{Network architecture.}
	\label{fig:architecture}
\end{figure*}

\begin{figure}[t]
	\centerline{\includegraphics[width=1\columnwidth]{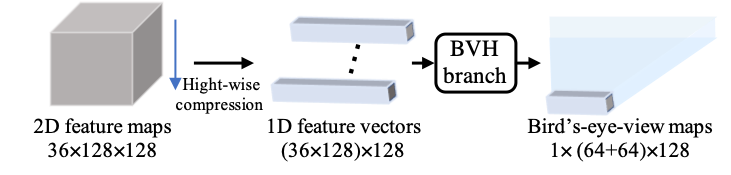}}
	\caption{Operations to convert the front-view features to a bird's eye view (shown in 3D camera space represented by 3D camera anchor maps).}
	\label{fig:bv_operations}
\end{figure}

\subsection{Network Architecture}

We develop a bird's-eye-view-based coarse-to-fine localization pipeline to estimate the 3D translation of all people in the scene in one shot. 
In Fig.~\ref{fig:architecture}, we present the network architecture of estimating five 2D maps and two 3D maps, which are used to generate the final results as shown in Fig.~2 of the main paper.
The input size $\vec{W}$ is $(512, 512)$. 
Following ROMP~\cite{romp}, we adopt a multi-head architecture and use HRNet-32~\cite{cheng2020higherhrnet} as backbone.
With backbone feature maps of size $\mathbb{R}^{32 \times H \times W}$, we employ three head branches to estimate four front-/bird's-eye-view 2D maps and a Mesh Feature map.

As illustrated in Fig.~\ref{fig:bv_operations}, our key design is to convert the front-view features to a bird's eye view via explicit operations including height-wise suppression and depth-wise exploration.
As shown in the middle branch of Fig.~\ref{fig:architecture}, we first explore the depth information of backbone features via a Bottleneck block.
And then we concatenate the explored depth features and front-view 2D maps as input to the BVH branch.
As shown in Fig.~\ref{fig:bv_operations}, we compress the 2D feature maps in height to obtain 1D feature vectors. 
In the BVH branch (Fig.~\ref{fig:architecture}), we employ six 1D convolution blocks to explicitly explore features in depth.
Two bird's-eye-view maps are of size $\mathbb{R}^{1 \times D \times W}$.

Next, we compose the front-view and bird's-eye-view maps to generate 3D maps.
We extend the front-view maps with an additional depth dimension and repeat $D$ times.
We also extend the bird's-eye-view maps with an additional height dimension and repeat $H$ times.
To obtain the 3D Center map, we multiply the bird's-eye-view Body Center heatmap to the front-view one and refine it with a 3D refiner (Fig.~\ref{fig:architecture}).
Then we add the bird's-eye-view offset map to the last channel of the front-view one to refine the depth.
To obtain the 3D Offset map, we further use a 3D refiner to refine the composed 3D maps, which improves the consistency between features of two views.

\subsection{Datasets}

In this section, we introduce the datasets we used during training and evaluation.

\textbf{AGORA~\cite{patel2021agora}} is a synthetic dataset with accurate annotations of body meshes and 3D translations, with 4,240 high-realism textured scans in diverse poses and clothes. 
Importantly, it contains 257 child scans. 
It contains 14K training and 3K test images. Each image has 5-15 people with frequent occlusions.
\textbf{AGORA-PC~\cite{patel2021agora}} is a high occlusion subset of the AGORA validation set. 
Each image has over 70\%  occlusion.
We use it to evaluate the performance under severe occlusion.
Note that there are no child samples in the validation set.

\textbf{Human3.6M~\cite{h36m}} is a single-person 3D pose dataset. 
It contains videos of 9 professional actors performing activities in 17 scenarios. 
It provides 3D pose annotations for each frame.
We sample every 5 frames to reduce redundancy.
We use its training set for training.

\textbf{MuCo-3DHP~\cite{mehta2018single}} is a synthetic multi-person 3D pose dataset. 
It is built on the single-person 3D pose dataset, MPI-INF-3DHP~\cite{mehta2018single}.
They use segmentation annotations to blend multiple single-person images into one.
For a fair comparison with 3DMPPE~\cite{moon2019camera}, we use the same synthetic version for training.

\textbf{Other 2D pose datasets.}
For better generalization, we also use four 2D pose datasets for training, including COCO~\cite{coco}, MPII~\cite{mpii}, LSP~\cite{lsp_extended}, and CrowdPose~\cite{crowdpose}. 
Besides, we also use the pseudo-3D annotations~\cite{joo2020eft} for training.

\subsection{Training Details}
The size of output maps are $H=W=128$, and $D=64$.
The threshold for the age offset is set to $t_{\alpha}=0.8$.
The FOV is set to $60^\circ$.
The loss weights are $w_{mpj}=200$, $w_{pmpj}=360$, $w_{pj2d}=400$, $w_{\theta}=80$, $w_{\beta}=60$, $w_{prior}=1.6$, $w_{cm}=100$, $w_{cm3d}=1000$, $w_{age}=4000$, and $w_{depth}=400$.
We train BEV on a server with four Tesla V100 GPUs.
The batch size is 64. The learning rate is 5$e^{-5}$.
The confidence threshold of the Body Center heatmap is 0.12.

Additionally, although we strive to alleviate the age bias in training samples, the age bias in existing 3D pose datasets is severe, and we have to use them to obtain good 3D pose estimation.
To handle the imbalanced distribution of the training sample space, we balance the sampling ratio of different datasets and evenly select the training samples from different age groups on RH.
The sampling ratios of different datasets are 16\% AGORA, 16\% MuCo-3DHP, 16\% RH, 18\% Human3.6M, 14\% COCO, 8\% CrowdPose, 6\% MPII, and 6\% LSP.

Also, we adopt a two-step training strategy. 
We first learn monocular 3D pose and shape estimation for 120 epochs on basic training datasets.
Then we add the weak annotations of RH to training samples and train for 120 epochs. 
If we need to fine-tune on AGORA, we add AGORA to the training sequence and train for 80 epochs.
In this process, the validation set of the RH is used to select checkpoints with good performance.

\subsection{Processing High-resolution Images}

As a one-stage method, BEV takes an image of constant size as input.
However, to process the high-resolution images, directly resizing them to a constant size would sacrifice the performance. 
Therefore, we develop a sliding window-based pipeline to achieve promising results on high-resolution images, as shown in Fig. 1 of the main paper.
In detail, we evenly split the image into multiple grids and then apply BEV on each grid.
This process is similar to the sliding window operation of 2D convolution. 
At each grid, we only take the result whose body center falls in the center area of the grid. 
Then we perform non-maximum suppression on the edge between grids to get rid of redundant predictions.
In this process, overlapping predictions with lower center confidence values will be deleted.

\section{Relative Human Dataset}

In this section, we provide more detailed analyses of our Relative Human dataset.

\begin{figure}[t]
	\centerline{\includegraphics[width=0.8\columnwidth]{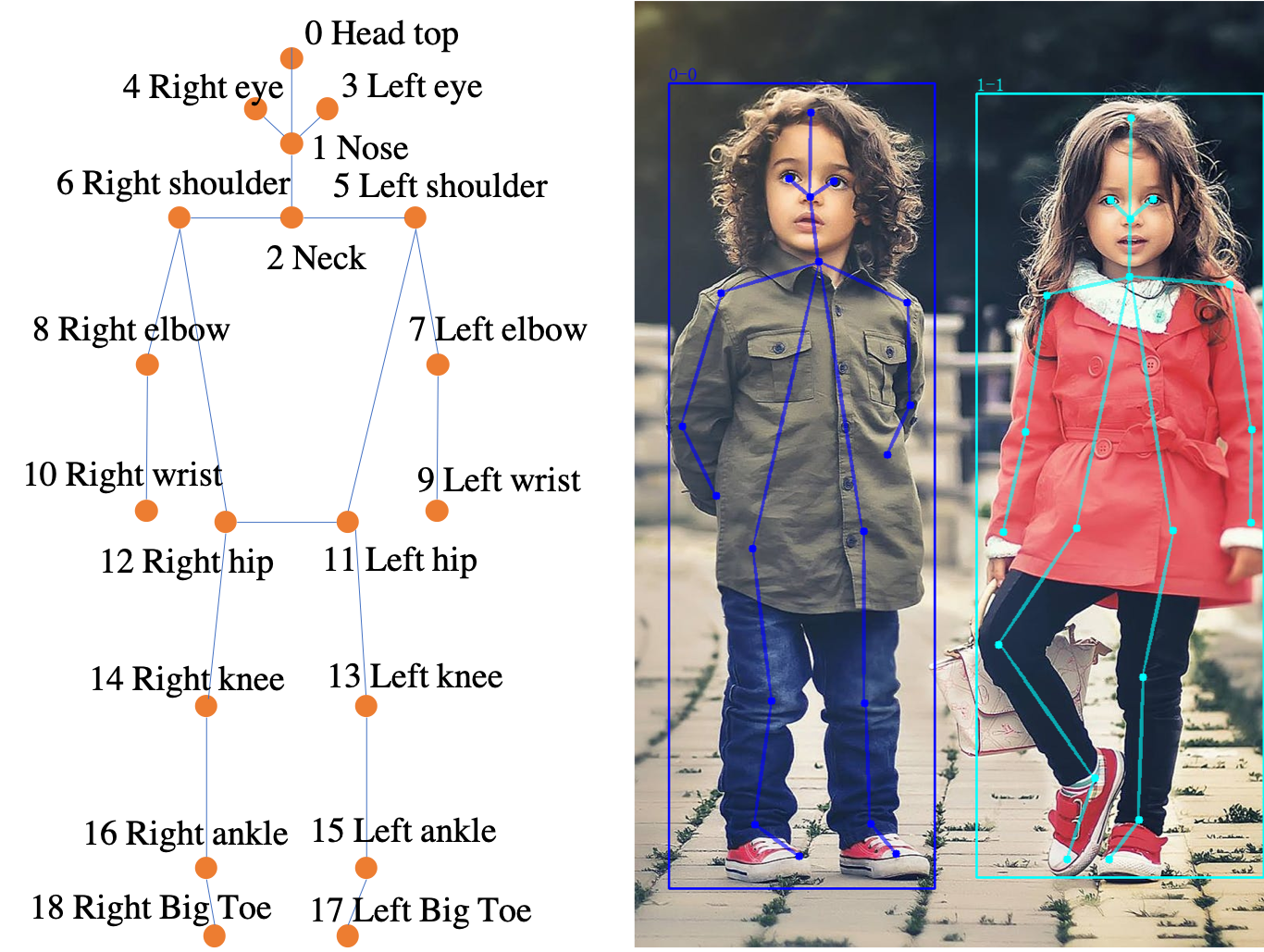}}
	\caption{The 2D skeleton definition.}
	\label{fig:RH_skeletons}
\end{figure}

In total, we collect about 7,689 images with weak annotations of 24,814 people.
We split them into three groups (5218, 635, 1836) for training, validation, and test respectively.
Among these images, about 1,000 images are collected from a free photo website~\cite{pexels} and we annotate the 2D poses defined as Fig.~\ref{fig:RH_skeletons}.
Note that compared with LSP's 14 keypoints, we add keypoints on the face and feet to represent their orientations.
The remaining images are selected from existing 2D pose datasets~\cite{coco,crowdpose,zhang2019pose2seg}.
We correct some erroneous 2D poses from the existing 2D pose dataset and add the missing detections.
Note that a large number of images in CrodPose~\cite{crowdpose} and OCHuman~\cite{zhang2019pose2seg} are selected from COCO~\cite{coco} and MPII~\cite{mpii}, which are also used as training samples by our compared methods~\cite{romp,jiang2020coherent,kolotouros2019spin,Kocabas_SPEC_2021}.
Therefore, we use these common images for training.

We classify all people in the image into four age groups, baby, child, teenager, and adult according to the following age ranges:
baby (0-3), kid (3-8),  teenager (8-16),  and adult (16+).
As shown in Tab.~\ref{tab:RH_subjects}, we provide the number of subjects in the four age groups and their proportions.
Compared with the existing multi-person 3D pose datasets~\cite{mehta2018single,3dpw,patel2021agora}, RH contains richer subjects and more occlusion cases.
Therefore, RH is more general and suitable for evaluating depth reasoning in the wild.

\textbf{The consistency of weak annotations.}
During the collection of weak annotations, we observe that people's judgments for such weak labels vary greatly.
It is hard to obtain consistent weak labels through online platforms (e.g. AMT).
Therefore, offline, we organized a group of labelers and trained them with unified standards.
To test how well they learn the standards, we prepare some pre-labeled data as test samples.
Ones who pass the test after training were employed for official labeling.
In addition, the annotations are double-checked by professional testers and the author.

\begin{table}[t]
\setlength\tabcolsep{0.5mm}
\small
	\centering
	\begin{tabular}{c|cccc}
	\toprule
RH splits & Babies & Children & Teenagers & Adults \\
\midrule
All & 1534 / 6\% & 2720 / 10\% & 1067 / 4\% & 19493 / 78\% \\
Train & 942 / 5\% & 1795 / 10\% & 690 / 4\% & 13478 / 79\% \\
Validation & 117 / 5\% & 209 / 9\% & 101 / 4\% & 1680 / 79\% \\
Test & 475 / 8\% & 716 / 12\% & 276 / 4\% & 4335 / 74\% \\
\midrule
\multicolumn{5}{c}{Existing multi-person 3D pose datasets}\\
\midrule
MuPoTS~\cite{mehta2018single} & - & - & - & 8 / 100\% \\
3DPW~\cite{3dpw} & - & - & - & 18 / 100\% \\
AGORA~\cite{patel2021agora} & - & 257 / 6\% & - & 3983 / 94\% \\
\bottomrule
\end{tabular}
	\caption{Subject number/proportions of four age groups on Relative Human (RH) and 3D pose benchmarks.}
	\label{tab:RH_subjects}
\end{table}

\begin{table*}[t]
	\footnotesize
	\centering
	\begin{tabular}{p{0.2cm}p{0.3cm}|l|ccc|cccc}
	\toprule
Input&& Method & \textbf{F1 score}$\uparrow$ & \textbf{Precision}$\uparrow$ & \textbf{Recall}$\uparrow$ & \textbf{MVE}$\downarrow$ & \textbf{MPJPE}$\downarrow$ & \textbf{NMVE}$\downarrow$ & \textbf{NMJE}$\downarrow$\\
\midrule
 \multirow{11}{*}{\rotatebox[origin=c]{90}{\textbf{Multi-stage}}} &
 \multirow{11}{*}{\rotatebox[origin=c]{90}{\textbf{Person crops from 3840x2160}}} &HMR~\cite{hmr} & 0.38 & 0.27 & 0.61 & 209.3 & 219.4 & 550.8 & 577.4 \\
&&SMPLify-X$\ddag$~\cite{smplx} & \textbf{0.57} & \textbf{0.60} & 0.55 & 213.3 & 208.3& 374.2& 365.4  \\
&&EFT~\cite{joo2020eft} & 0.43 & 0.34 & 0.60 & 193.5 & 202.7 & 450.0 & 471.4   \\
&&SPIN~\cite{kolotouros2019spin} & 0.33 & 0.23 & 0.61 & 193.2 & 203.7 & 585.5 & 617.3 \\
&&ExPose~\cite{expose}& 0.53 & 0.46 & 0.61 & 174.0 & 176.6 & 328.3 & 333.2 \\
&&Frankmocap~\cite{rong2021frankmocap} & 0.40 & 0.30 & 0.62 & 204.2 & 203.7 & 510.5 & 509.2 \\
&&PyMAF~\cite{zhang2021pymaf}& 0.27 & 0.16 & 0.82 & 192.0 & 203.2 & 711.1 & 752.6  \\
&&PIXIE~\cite{feng2021pixie} & 0.48 & 0.39 & 0.61 & 174.6 & 174.7 & 363.8 & 364.0\\
&&SPIN$^\star$~\cite{patel2021agora}& 0.31 & 0.21 & 0.60 & 186.7 & 191.7 & 602.3 & 618.4  \\
&&SPEC$^\star$~\cite{Kocabas_SPEC_2021} & 0.52 & 0.40 & 0.73 & 163.2 & 171.0 & 313.8 & 328.8 \\
&&PARE~\cite{kocabas2021pare} & 0.55 & 0.44 & 0.74 & 186.4 & 193.9 & 338.9 & 352.5\\
&&Pose2Pose$^\star$$^\dagger$~\cite{moon2020pose2pose}& 0.56 & 0.40 & \textbf{0.91} & \textbf{146.4}& \textbf{153.3} &\textbf{261.4} & \textbf{273.8} \\
\midrule
\multirow{5}{*}{\rotatebox[origin=c]{90}{\textbf{One-stage}}} & \multirow{5}{*}{\rotatebox[origin=c]{90}{\textbf{512x512}}} &ROMP~\cite{romp}& 0.38 & 0.39 & 0.37 & 198.5 & 207.4 & 522.4 & 545.8 \\
&&BEV w/o WST & 0.41 & 0.39 & 0.45 & 194.4 & 202.6 & 474.1 & 494.1  \\
&&ROMP$^\star$~\cite{romp} & 0.50 & 0.37 & 0.80 & 156.6 & 159.8 & 313.2& 319.6 \\
&&BEV$^\star$ w/o WST  & \textbf{0.58} & \textbf{0.44} & \textbf{0.86} & 146.0 & 148.3 & 251.7 & 255.7\\
&&BEV$^\star$ & 0.55 & 0.41 & 0.85 & \textbf{125.9} &\textbf{ 129.1} & \textbf{228.9} & \textbf{234.7} \\
\bottomrule
\end{tabular}
	\caption{Comparison to existing SOTA methods on the ``AGORA kids" test set. Results are obtained from the AGORA leaderboard. $^\star$ is fine-tuning on the AGORA training set or synthetic data~\cite{Kocabas_SPEC_2021} generated in the same way as AGORA. $^\ddag$ means the optimization-based method while the rest are learning-based methods.  $^\dagger$ means the paper is under review.}
	\label{tab:agora_kid}
\end{table*}

\begin{table*}[t]
	\footnotesize
	\centering
	\begin{tabular}{p{0.2cm}p{0.3cm}|l|ccc|cccc}
	\toprule
Input&& Method & \textbf{F1 score}$\uparrow$ & \textbf{Precision}$\uparrow$ & \textbf{Recall}$\uparrow$ & \textbf{MVE}$\downarrow$ & \textbf{MPJPE}$\downarrow$ & \textbf{NMVE}$\downarrow$ & \textbf{NMJE}$\downarrow$\\
\midrule
\multirow{11}{*}{\rotatebox[origin=c]{90}{\textbf{Multi-stage}}} &
 \multirow{11}{*}{\rotatebox[origin=c]{90}{\textbf{Person crops from 3840x2160}}} &HMR~\cite{hmr} & 0.80 & 0.93 & 0.70 & 173.6 & 180.5 & 217.0 & 226.0 \\
&&SMPLify-X $^\ddag$~\cite{smplx} & 0.71 & 0.86 & 0.60 & 187.0 & 182.1 & 263.3 & 256.5 \\
&&EFT~\cite{joo2020eft} & 0.69 & \textbf{0.97} & 0.54 & 159.0 & 165.4 & 196.3 &  203.6  \\
&&SPIN~\cite{kolotouros2019spin} & 0.78 & 0.91 & 0.69 &  168.7 & 175.1 & 216.3 & 223.1 \\
&&ExPose~\cite{expose}& 0.82 & 0.96 & 0.71 & 151.5 & 150.4 & 184.8 & 183.4  \\
&&Frankmocap~\cite{rong2021frankmocap} & 0.80 & 0.93 & 0.71 & 204.2 & 203.7 & 510.5 & 509.2 \\
&&PyMAF~\cite{zhang2021pymaf}& 0.84 & 0.86 & 0.82 & 192.0 & 203.2 & 711.1 & 752.6  \\
&&PIXIE~\cite{feng2021pixie} & 0.82 & 0.95 & 0.73 & 142.2 & 140.3 & 173.4 & 171.1 \\
&&SPIN$^\star$~\cite{patel2021agora}& 0.77 & 0.91 & 0.67 & 168.7 & 175.1 & 216.3 & 223.1\\
&&SPEC$^\star$~\cite{Kocabas_SPEC_2021} & 0.84 & 0.96 & 0.74 & 106.5 & 112.3 & 126.8 & 133.7 \\
&&PARE~\cite{kocabas2021pare} & 0.84 & 0.96 & 0.75 & 140.9 & 146.2 & 167.7 & 174.0\\
&&Pose2Pose$^\star$$^\dagger$~\cite{moon2020pose2pose}& \textbf{0.94} & 0.94 & \textbf{0.93} & \textbf{84.8} & \textbf{89.8} & \textbf{90.2} & \textbf{95.5} \\
\midrule
\multirow{5}{*}{\rotatebox[origin=c]{90}{\textbf{One-stage}}} & \multirow{5}{*}{\rotatebox[origin=c]{90}{\textbf{512x512}}}&ROMP~\cite{romp}& 0.69 & 0.97 & 0.54 & 161.4 & 168.1 & 233.9 & 242.3 \\
&&BEV w/o WST & 0.75 & \textbf{0.97} & 0.61 & 164.2 & 169.1 & 218.9 & 225.5\\
&&ROMP$^\star$~\cite{romp} & 0.91 & 0.95 & 0.88 & 103.4 & 108.1 & 113.6 & 118.8 \\
&&BEV$^\star$ w/o WST & 0.93 & 0.96 & 0.90 & 105.6 & 109.7 & 113.5 & 118.0\\
&&BEV$^\star$  & \textbf{0.93} & 0.96 & \textbf{0.90} & \textbf{100.7} & \textbf{105.3} & \textbf{108.3} & \textbf{113.2} \\
\bottomrule
\end{tabular}
	\caption{Comparison to existing SOTA methods on AGORA full test set. Results are obtained from the AGORA leaderboard. $^\star$ is fine-tuning on the AGORA training set or synthetic data~\cite{Kocabas_SPEC_2021} generated in the same way as AGORA. $^\ddag$ means the optimization-based method while the rest are learning-based methods. $^\dagger$ means the paper is under review.}
	\label{tab:agora_full}
\end{table*}

\section{Discussion}

\textbf{Why not estimate the 3D heatmap directly?}
The main challenge is the lack of sufficient multi-person data with accurate 3D translation annotations for supervision, especially for in-the-wild cases. 
Due to the data lack problem, directly learning 3D heatmap performs poorly.
It is hard to effectively supervise multi-person 3D heatmaps with 2D annotations.
In contrast, our separable representation disentangles the 3D heatmap into the front-view and the bird's-eye-view maps. 
In this way, our model can learn robust front-view localization from abundant 2D in-the-wild datasets. 
With robust front-view attention, the model can focus on learning depth reasoning from weak annotations in RH with the proposed WST.

\textbf{Heatmap refinement and decomposition.}
Following previous methods~\cite{romp}, we also adopt the powerful heatmap representation for detection. While its rough granularity limits its effectiveness in fine localization. 
Some previous methods explore the refinement and decomposition of the heatmap to alleviate this problem.
PifPaf~\cite{kreiss2019pifpaf} estimates offset maps to refine the coarse 2D pose coordinates parsed from the heatmap. 
VNect~\cite{mehta2017vnect} estimates three 2D maps containing x/y/z coordinates of the 3D pose at each position.
Luvizon et al.~\cite{luvizon20182d} employ soft-argmax to decompose 2D/3D heatmap into 1D for separate supervision, while it does not deal with multiple overlapping people.
Different from previous solutions, we propose a novel bird's-eye-view-based representation for multi-person 3D localization.
As we introduced above, it disentangles the depth-wise information into an individual map for easier learning. 
We also estimate a 3D offset map to improve the granularity of 3D localization. 

\section{Quantitative and Qualitative Results}

In this section, we first show more comparisons to SOTA methods on AGORA and then provide more qualitative results on Internet images, CMU Panpotic~\cite{cmu_panoptic}, AGORA~\cite{patel2021agora}, and RH.

\begin{figure*}[t]
	\centerline{\includegraphics[width=1.00\textwidth]{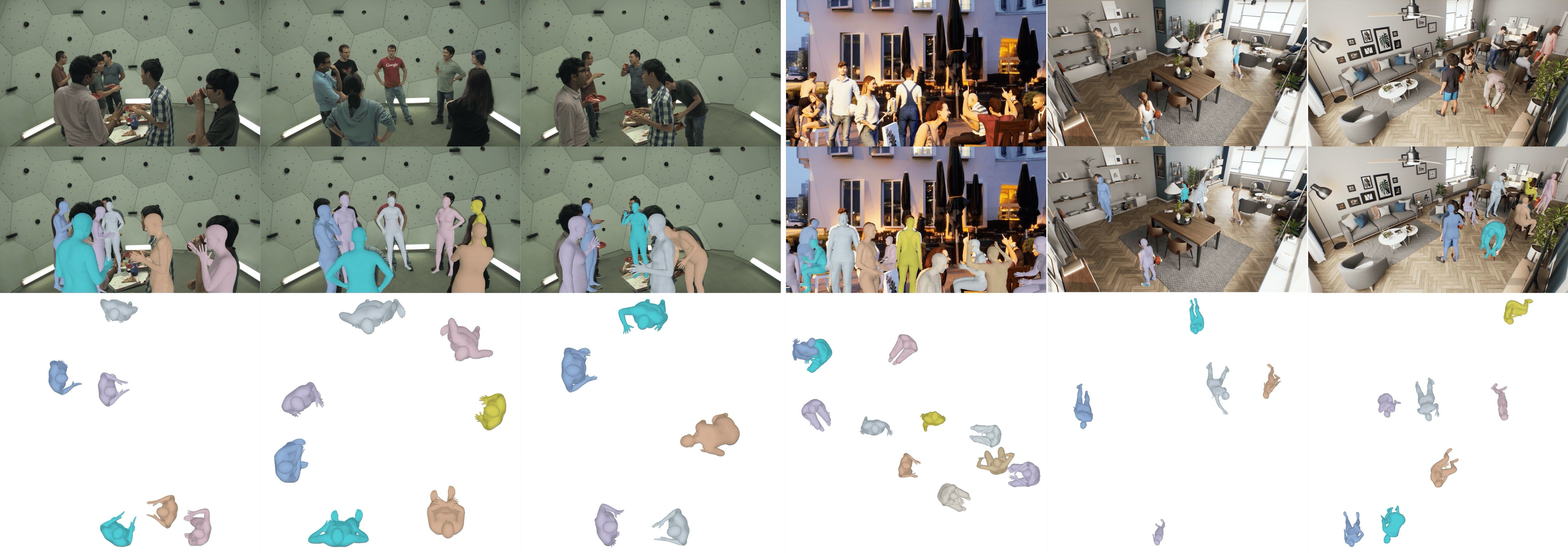}}
	\caption{Qualitative results on CMU Panoptic~\cite{cmu_panoptic} and AGORA~\cite{patel2021agora} datasets.}
	\label{fig:CMUP_AGORA}
\end{figure*}

\begin{figure*}[t]
	\centerline{\includegraphics[width=0.96\textwidth]{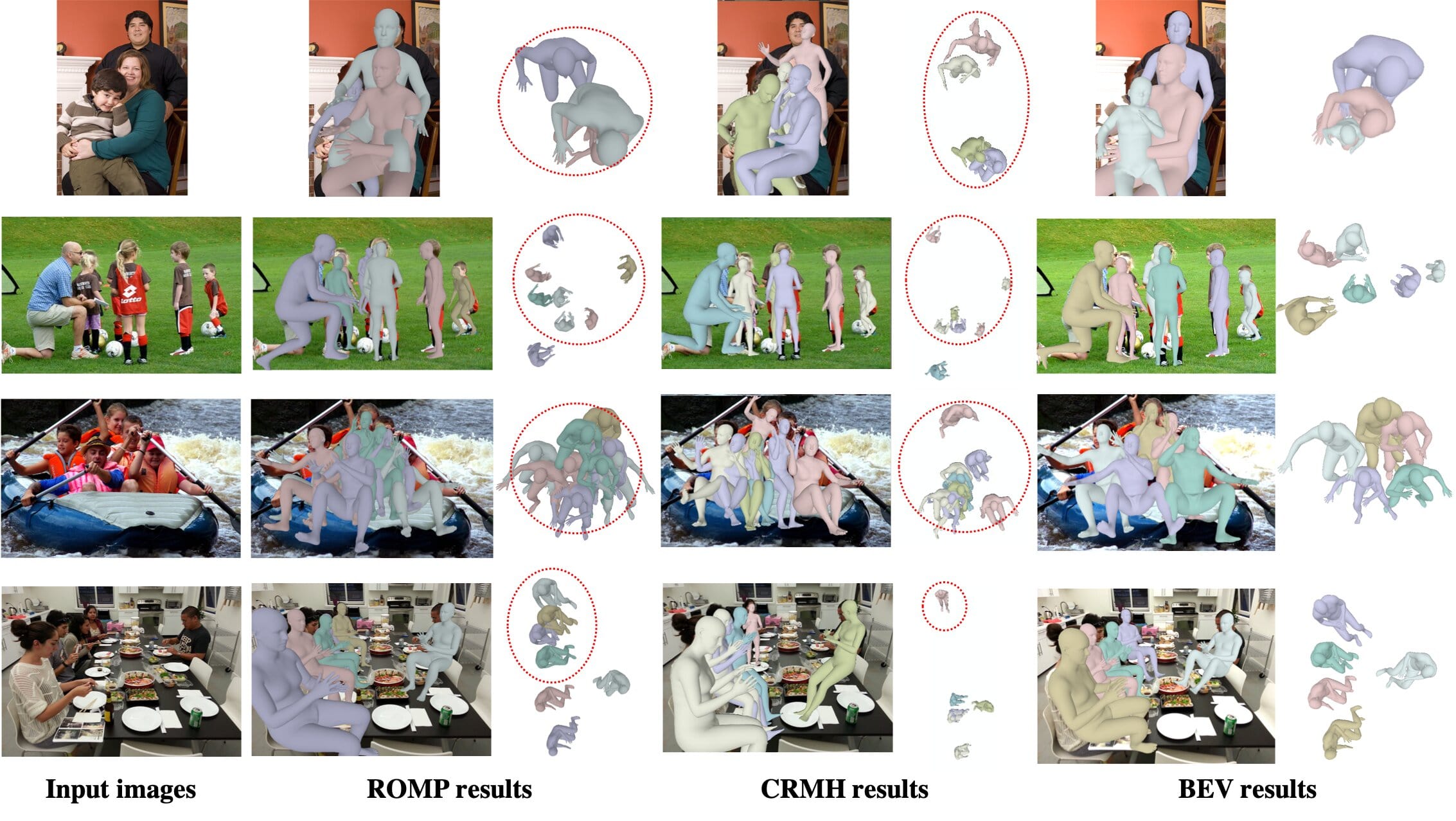}}
	\caption{Qualitative comparisons to SOTA methods, ROMP~\cite{romp} and CRMH~\cite{jiang2020coherent}, on RH test set.}
	\label{fig:qualitative_comparisons_RH_test}
\end{figure*}

\subsection{Quantitative Comparisons}
In Tab.~\ref{tab:agora_kid} and~\ref{tab:agora_full}, we show the results of existing SOTA methods on ``AGORA kids'' and the full test set respectively.
Results in Tab.~\ref{tab:agora_kid} show that BEV outperforms all previous methods by a large margin in terms of child mesh reconstruction.
It demonstrates that learning weak annotations via the proposed weakly-supervised training (WST) helps to alleviate the age bias.
Multi-stage methods, like Pose2Pose~\cite{moon2020pose2pose}, benefit from taking high-resolution person crops as input, which helps process the small-scale subjects in AGORA.
Besides, as a sanity check, we also compare with SOTAs on 3DPW and MuPoTS datasets.
While not tuned for uncrowded scenes, BEV is on par with the previous methods on MuPoTs (Tab.~\ref{tab:MuPoTS}) and 3DPW (Tab.~\ref{tab:3DPW}).

\begin{table}[t]
  \centering
    \begin{tabular}{l|cccc|c|cc}
    \toprule
    Method &  All$\uparrow$ & Matched$\uparrow$\\
   \midrule
        CRMH~\cite{jiang2020coherent} & 69.1 & 72.2 \\
        ROMP~\cite{romp} & 69.9 & 74.6 \\
        3DCrowdNet~\cite{choi20223dcrowdnet} & \textbf{72.7} & 73.3 \\
        BEV & 70.2 & \textbf{75.2} \\
   \bottomrule
    \end{tabular}
    \caption{Comparisons to the SOTAs on MuPoTS. }\label{tab:MuPoTS}
\end{table}

\begin{table}[t]
	\centering
	\begin{tabular}{l|ccc}
	\toprule
{Method} & {PMPJPE} & {MPJPE}& {MPVE} \\
\midrule
HybrIK~\cite{li2021hybrik} & 48.8 & 80.0 & 94.5 \\
METRO~\cite{lin2021end} & 47.9 & 77.1 & \textbf{88.2} \\
ROMP~\cite{romp} & 47.3 & \textbf{76.7} & 93.4 \\
BEV & \textbf{46.9} & 78.5 & 92.3 \\
\bottomrule
\end{tabular}
	\caption{Comparisons to the SOTAs on 3DPW test set.}
	\label{tab:3DPW}
\end{table}

\subsection{Ablation Studies}

\begin{table}[t]
	\footnotesize
	\setlength\tabcolsep{4pt}
	\centering
	\begin{tabular}{l|c|cccc}
	\toprule
Method & F1 score$\uparrow$ & MVE$\downarrow$ & MPJPE$\downarrow$ & NMVE$\downarrow$ & NMJE$\downarrow$\\
\midrule
ROMP~\cite{romp} & 0.695 &173.76 & 170.55 & 249.96 & 245.34\\
BEV & \textbf{0.732} & \textbf{169.21} & \textbf{165.27} & \textbf{231.16} & \textbf{225.76}\\
\midrule
w/o WST & 0.738 & 171.16 &168.12 & 235.06 & 230.89\\
w/o DC & 0.741 & 170.59 & 168.12 & 229.98 & 225.67 \\
\bottomrule
\end{tabular}
	\caption{3D mesh/pose error on AGORA-PC, the high occlusion (over 70\%) subset of the AGORA validation set (no kids). }
	\label{tab:AGORA_PC}
\end{table}

To analyse the performance gain of different designs, we perform more ablation studies on AGORA$-$PC, a high occlusion (over 70\%) subset of the AGORA validation set (no kids).
This has ground truth 3D annotations for detailed evaluation while the test set does not. BEV uses the same training samples as~\cite{romp}. 
Comparing BEV and BEV w/o WST in Tab.~\ref{tab:AGORA_PC} also shows that our gains in high occlusion situations come from the 3D representation.

Besides, we also evaluate the effectiveness of depth encoding (DC) for 3D mesh parameter regression.
Depth encoding is developed to transfer people at different depths to individual feature spaces.
Tab.~\ref{tab:AGORA_PC} shows that adding the depth encoding reduces mesh reconstruction error under high occlusion (over 70\%).
It demonstrates that achieving depth-aware mesh regression via adding depth encoding is beneficial to alleviating depth ambiguity and improving the stability under occlusion.

\subsection{Qualitative Results}
\begin{figure*}[t]
	\centerline{\includegraphics[width=1.00\textwidth]{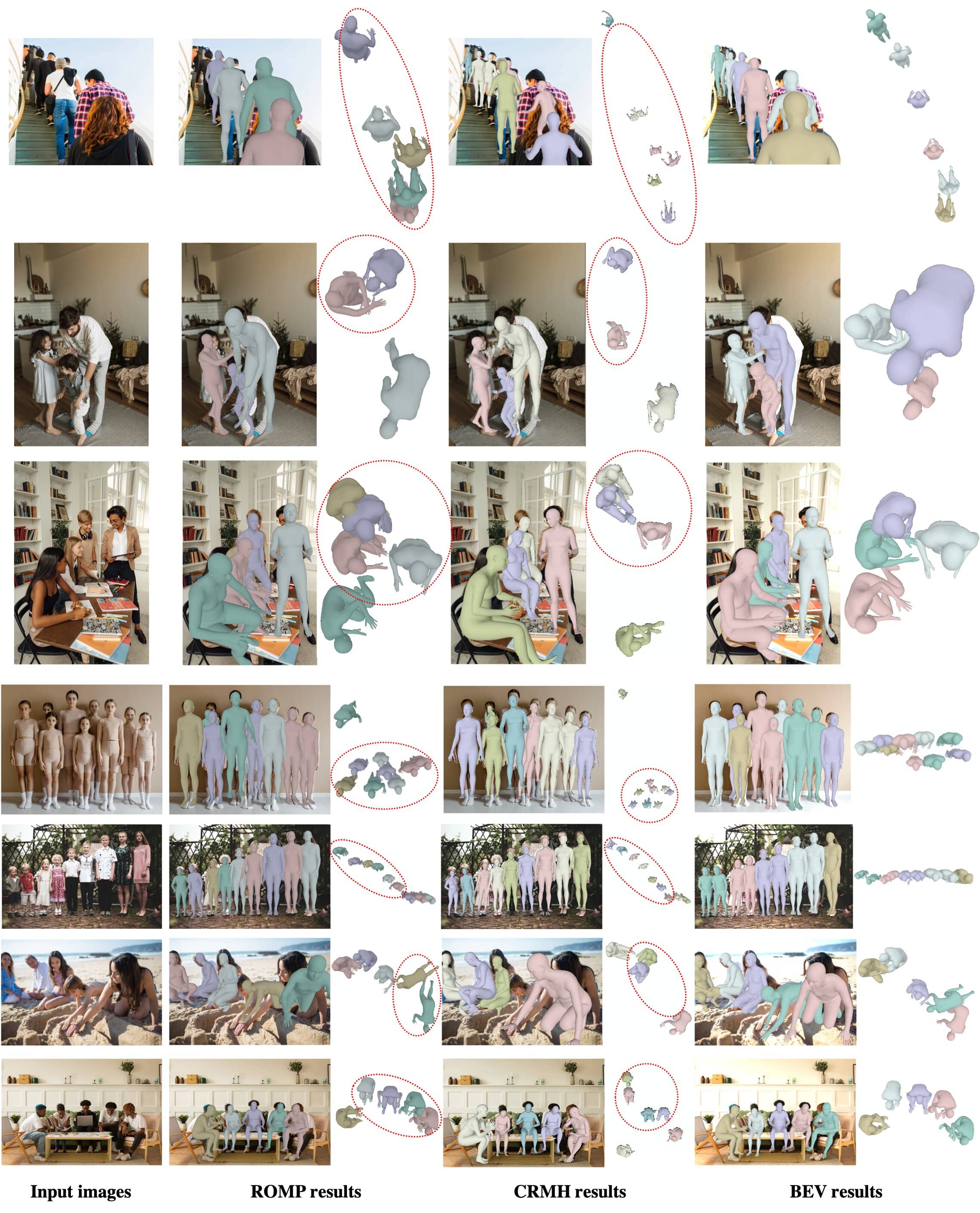}}
	\caption{Qualitative comparisons to SOTA methods, ROMP~\cite{romp} and CRMH~\cite{jiang2020coherent}, on Internet images~\cite{pexels}.}
	\label{fig:qualitative_comparisons_pexels}
\end{figure*}

\begin{figure*}[t]
	\centerline{\includegraphics[height=0.96\textheight]{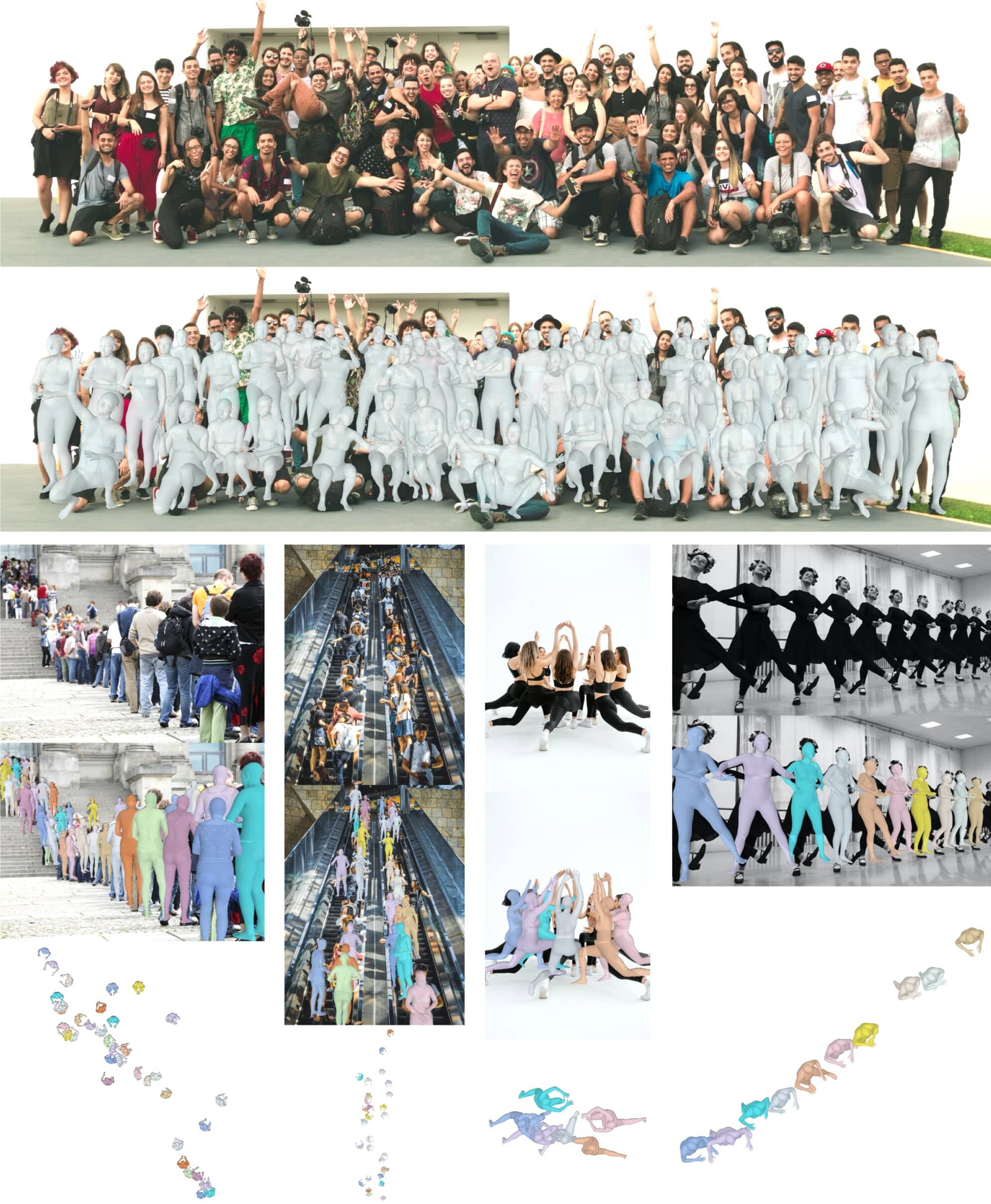}}
	\caption{Failure cases on Internet images~\cite{pexels}}
	\label{fig:failure_cases}
\end{figure*}

In Fig.~\ref{fig:CMUP_AGORA}, we present more qualitative results on CMU Panoptic and AGORA. 
In Fig.~\ref{fig:internet},~\ref{fig:qualitative_comparisons_RH_test},~\ref{fig:qualitative_comparisons_pexels}, we show the results under various crowded scenarios, including queuing, standing side by side, and mixed scenarios.
Compared with ROMP~\cite{romp} and CRMH~\cite{jiang2020coherent}, BEV performs much better in detection, depth reasoning, and robustness to  occlusion, especially in cases containing children.
These results demonstrate the superiority of our 3D representation, WST, and perspective camera model. 
However, we also observe some limitations of BEV from failure cases in Fig.~\ref{fig:failure_cases}.
Without modeling the contact between multiple people, BEV may miss obvious contact and cannot avoid mesh intersections.
Besides, BEV is unable to handle occlusions with few visible parts and dense small-scale subjects in crowds.

{\small
\bibliographystyle{ieee_fullname}
\bibliography{suppmat}
}